\documentclass[a4paper,fleqn]{cas-dc}

\usepackage[authoryear,longnamesfirst]{natbib}
\usepackage{subfig}
\def\tsc#1{\csdef{#1}{\textsc{\lowercase{#1}}\xspace}}
\tsc{WGM}
\tsc{QE}
\begin{document}
\let\WriteBookmarks\relax
\def\floatpagepagefraction{1}
\def\textpagefraction{.001}

\shorttitle{Vision and Language Integration for Domain Generalization}    

\shortauthors{Wang Yanmei et~al.}  

\title [mode = title]{Vision and Language Integration for Domain Generalization}

\affiliation[1]{organization={State Key Laboratory of Robotics, Shenyang Institute of Automation, Chinese Academy of Sciences},
	%         addressline={}, 
	city={Shenyang},
	%          citysep={}, % Uncomment if no comma needed between city and postcode
	postcode={110016}, 
	country={China}}
\affiliation[2]{organization={Institutes for Robotics and Intelligent Manufacturing, Chinese Academy of Sciences},
	%         addressline={}, 
	city={Shenyang},
	%          citysep={}, % Uncomment if no comma needed between city and postcode
	postcode={110169}, 
	country={China}}
\affiliation[3]{organization={University of Chinese Academy of Sciences},
	%         addressline={}, 
	city={Beijing},
	%          citysep={}, % Uncomment if no comma needed between city and postcode
	postcode={100049}, 
	country={China}}

\author[1,2,3]{Yanmei Wang}[]

\author[1,2]{Xiyao Liu}[]
\author[1,2,3]{Fupeng Chu}[]
%\ead{wangyanmei@sia.cn}
\fnmark[1]
\author[1,2]{Zhi Han}[]
\ead{hanzhi@sia.cn}
\ead[url]{https://orcid.org/0000-0002-8039-6679}
\cormark[1]


%\affiliation[<aff no>]{organization={},
	%%          citysep={}, % Uncomment if no comma needed between city and postcode

\cortext[1]{Corresponding author}

\begin{abstract}
Domain generalization aims at training on source domains to uncover a domain-invariant feature space, allowing the model to perform robust generalization ability on unknown target domains. However, due to domain gaps, it is hard to find reliable common image feature space, and the reason for that is the lack of suitable basic units for images.
 Different from image in vision space, language has comprehensive expression elements that can effectively convey semantics. 
Inspired by the semantic completeness of language and intuitiveness of image, we propose VLCA, which combine language space and vision space, and connect the multiple image domains by using semantic space as the bridge domain. Specifically, in language space, by taking advantage of the completeness of language basic units, we tend to capture the semantic representation of the relations between categories through word vector distance. Then, in vision space, by taking advantage of the intuitiveness of image features, the common pattern of sample features with the same class is explored through low-rank approximation. In the end, the language representation is aligned with the vision representation through the multimodal space of text and image. Experiments demonstrate the effectiveness of the proposed method. 
\end{abstract}


% \begin{highlights}
% \item Combining the advantages of completeness of language and intuitiveness of image features, multimodal semantic space is used as the bridge domain to propose a domain generalization method based on language and vision space.
% \item In multimodal space, the domain embedding of the domain prompt is constructed  by CLIP, and then  domain embeddings are constrained to be orthogonal to the image features, thereby reducing the influence of the domain variations.
% \item In the language space, the semantic distance is measured between classes through word vectors to provide interclass relationship supervision for network training.
% \item In vision space, low-rank decomposition is performed on the features matrix of the same category,  and the features matrix is made approximate the subspace spanned by the largest singular value, so as to obtain common patterns of the same class samples.
% \end{highlights}

% Keywords
\begin{keywords}
 Domain generalization\sep vision language model\sep low-rank matrix\sep word vector.
\end{keywords}

\maketitle

\section{Introduction}
D{eep} neural networks have demonstrated significant performance in many applications \cite{10461100, 10452765, 10366875, wu2021multi, wu2019mutually, jia2024causal, li2024ewt, lin2024multi, wu2019cross, zhu2020asta, liu2024segmenting}. However, the training of neural networks is influenced by the distribution of the training data. Specifically, due to the fact that the number of parameters in deep neural networks far exceeds the number of samples, the network model tends to overfit the distribution of the training data, which leads to learning features that are specific to the distribution of the learned data. As a result, the model may perform poorly on data from other distributions. To address the aforementioned issues, domain generalization has attracted increasing attention. The aim of domain generalization is to train the network on source domains, enabling it to generalize effectively on unseen target domains.
\begin{figure}[!h]
    \centering
	\includegraphics[width=0.5\textwidth]{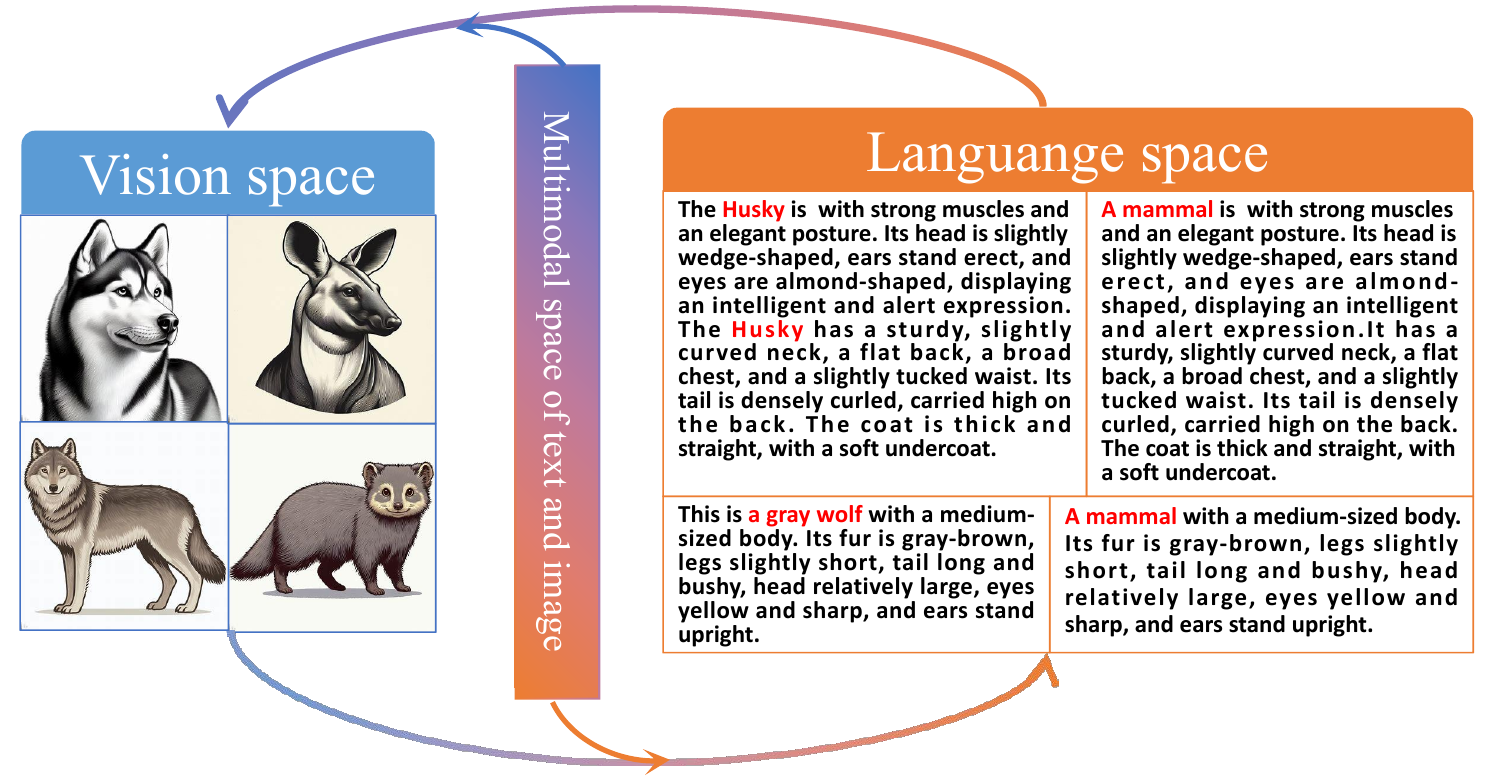}
    \caption{The ambiguity of the text descriptions in language space and the intuitiveness of images in vision space. The descriptions of Husky and gray wolf are generated by GPT-3.5 \cite{floridi2020gpt}. The descriptions of the two are approximate. Then, we use the DALL$\cdot$E \cite{reddy2021dall} to generate the picture of Husky and gray wolf according to the descriptions, as shown in the two pictures of the first column. In the second column, the ``Husky" and ``gray wolf" in the text descriptions are replaced with ``mammal", and the other descriptions remain unchanged, resulting in two other animals that fit the description. We attempt to combine language space and vision space and connect the multiple image domains by using semantic space as the bridge domain.}
	\label{wolfanddog}
\end{figure}

The basic units in natural language processing (NLP) are words. Similar to words in language, the basic units in images can be seen as pixels. However, unlike words in language, pixels in the image do not possess explicit semantic concepts. They solely represent color information at a specific location within the image. Training a convolutional neural network (CNN) allows us to obtain a set of filters that serve as effective feature extractors. Extracted features can express the semantic information of an image. However, as mentioned earlier, neural networks can easily overfit the distribution of the training data when extracting features. This may lead to the extraction of domain-specific features, making it challenging to effectively capture features from other domains. For instance, image features such as color and texture extracted from the \textit{``photo"} domain may not be applicable to the \textit{``sketch"} domain, which primarily consists of simple outlines and strokes.

With the emergence and development of word vector \cite{pennington2014glove, church2017word2vec} and CLIP \cite{radford2021learning}, applying semantic supervision from NLP to image processing has become possible. However, language is ambiguous. For instance, for someone who has never seen a dog or a wolf, we cannot effectively convey the characteristic differences between the two using language alone, as shown in Fig. \ref{wolfanddog}. The descriptions of Husky and gray wolf are generated by GPT-3.5 \cite{floridi2020gpt} with the prompt \textit{``What dose a Husky look like?"}. Then, we use the DALL$\cdot$E \cite{reddy2021dall} to generate the picture of Husky and gray wolf according to the descriptions, as shown in the two pictures of the first column. However, when the ``Husky" and ``gray wolf" in the text descriptions are replaced with ``mammal", and the other descriptions remain unchanged, it results in two other animals in the second column, which also fit the descriptions. In contrast, images as a more advanced expression form than language, provide a more intuitive way to represent features. 

Inspired by the above observations, to fully leverage the semantic completeness of language basic units and the intuitiveness of image features, we attempt to integrate the basic units of language with the explicit representation in vision, which means using semantic space as the bridge domain to connect multiple image domains so as to learn domain-invariant features. In language space, word vectors are used to supervise the relative relationships between categories. In vision space, the common patterns among samples of the same category are explored through matrix low-rank approximation. Following these, the multimodal space of text and images serves as a bridge to connect the language space and vision space. This is achieved by constraining the domain-specific information, making domain embedding and image features orthogonal.
%In the semantic space, constrainting image features and domain embedding are orthogonal. The domain embeddings are generated by CLIP according to the domian prompts. Thereby weaken the influence of domain information in the image features. In addation, word vectors are used to supervise the relative relationships between categories. In the image space, the common patterns of among samples of the same category are explored through matrix low-rank decomposition.

The contributions of the method can be summarized as follows:
\begin{itemize}
	\item Combining the advantages of completeness of language and intuitiveness of image features, multimodal semantic space is used as the bridge domain to propose a domain generalization method based on language and vision space.
	\item In multimodal space, the domain embedding of the domain prompt is constructed  by CLIP, and then  domain embeddings are constrained to be orthogonal to the image features, thereby reducing the influence of the domain variations.
	\item In the language space, the semantic distance is measured between classes through word vectors to provide interclass relationship supervision for network training.
	\item In vision space, low-rank decomposition is performed on the features matrix of the same category,  and the features matrix is made approximate the subspace spanned by the largest singular value, so as to obtain common patterns of the same class samples.
	%	\item The low-rank decomposition of the feature matrix is applied to the image feature space to obtain the common feature patterns of the same class samples. 
	%	\item The feature matrix composed of the same category samples  is projected into the space composed of left singular vectors, so as to find the common feature patterns of samples of the same class in different domains
\end{itemize}

\section{Related Work}
\label{sec:related work}
\subsection{Domain Generalization}
Domain generalization aims to address the performance degradation of a model when faces new, unseen domains or environments. In the real world, models are often trained on a specific dataset or environment, however, when the model is applied to a domain different from the training data, its performance can degrade significantly. There are many studies on domain generalization methods, including sample augmentation \cite{ chaari2018frequency, xu2021fourier, kang2022style, li2024takes}, feature selection techniques \cite{muandet2013domain, li2018domain, dou2019domain, liu2021domain, niu2023knowledge, zhou2023value, ng2024improving}, and meta-learning methods \cite{li2018learning, balaji2018metareg, du2020learning}. Feature selection helps the model learn features that are common to all domains, reducing dependence on specific domains. Domain adaptation technology aims to effectively transfer knowledge between different domains by adjusting the parameters of the model. The meta-learning method simulates the situation in different domains during the training process to enhance the model's adaptability to unknown domains. Different from the above method, we use joint vision-language space to improve the generalization performance of models in unknown domains.
\subsection{Vision-Language Models}
Vision-Language Models combines models for image and natural language processing to enable understanding and interaction between images and text \cite{dusurvey}. The purpose is to enable computers to understand and process both image and text data. This type of model consists of two main components: the Visual Processing Module and the Language Processing Module. Integrating visual and linguistic information enables the model to understand the semantic relationship between image and text. There are many studies on visual language pre-training models. For example, \cite{radford2021learning} proposes CLIP, a contrast-based learning method that enables a joint understanding of images and text by mapping images and text into a shared embedded space so that similar images and text are closer together in that space. \cite{reddy2021dall} is able to generate relevant images from natural language descriptions. It combines text description with image generation, providing users with a new, text-based way of image generation.
\subsection{Word Vector}
Word vector is a technique in natural language processing (NLP) that maps words to real vectors. It represents words in a continuous way, allowing for better processing of textual information for the computer. 
%Word vectors map words into a continuous real vector space, allowing the semantic information of words to be represented in a distributed manner. 
In vector space, semantically similar words are mapped to nearby locations, so that relationships between words can be better captured. GloVe \cite{pennington2014glove} and Word2Vec \cite{church2017word2vec} are two commonly used word vector pre-training models. Both are trained on large-scale corpus and provide pre-trained word vectors. In the paper, we use word vectors to represent the inter-class relationship of category labels.
\subsection{Low-rank Matrix Decomposition}
%Low-rank matrix factorization employs the inherent structural similarities among matrix elements to filter out noise and eliminate redundant information. By decomposing the matrix into a low-rank component and a sparse component, RPCA \cite{candes2011robust, wright2009robust, sun2013robust} stands out as an effective matrix factorization method for extracting crucial insights from data. The low-rank component captures the dominant underlying patterns or features present in the data, providing a concise representation that encapsulates the essential information. And it effectively serves as a denoising mechanism, enabling the extraction of meaningful and robust patterns \cite{sun2013robust}.
Matrix low-rank decomposition is a widely used technique in data dimensional reduction, feature extraction, and noise removal \cite{feng2017robust, liu2011latent, chen2016nonconvex}. Traditional methods such as principal component analysis (PCA) \cite{mackiewicz1993principal} and singular value decomposition (SVD) \cite{hoecker1996svd} have become commonly used tools to efficiently extract the main feature information of the data. SVD decomposition involves breaking down a matrix into the product of three matrices: the left singular matrix, the singular values, and the right singular matrix. 
%The left singular matrix represents the shared patterns among the row vectors of the matrix, while the right singular vectors represent the shared patterns among the column vectors of the matrix. 
In this paper, we form a matrix with features from samples of the same category and employ SVD to extract common patterns among the samples.
\section{Method}
\begin{figure*}[!h]
	\centering
	\includegraphics[width=1\textwidth]{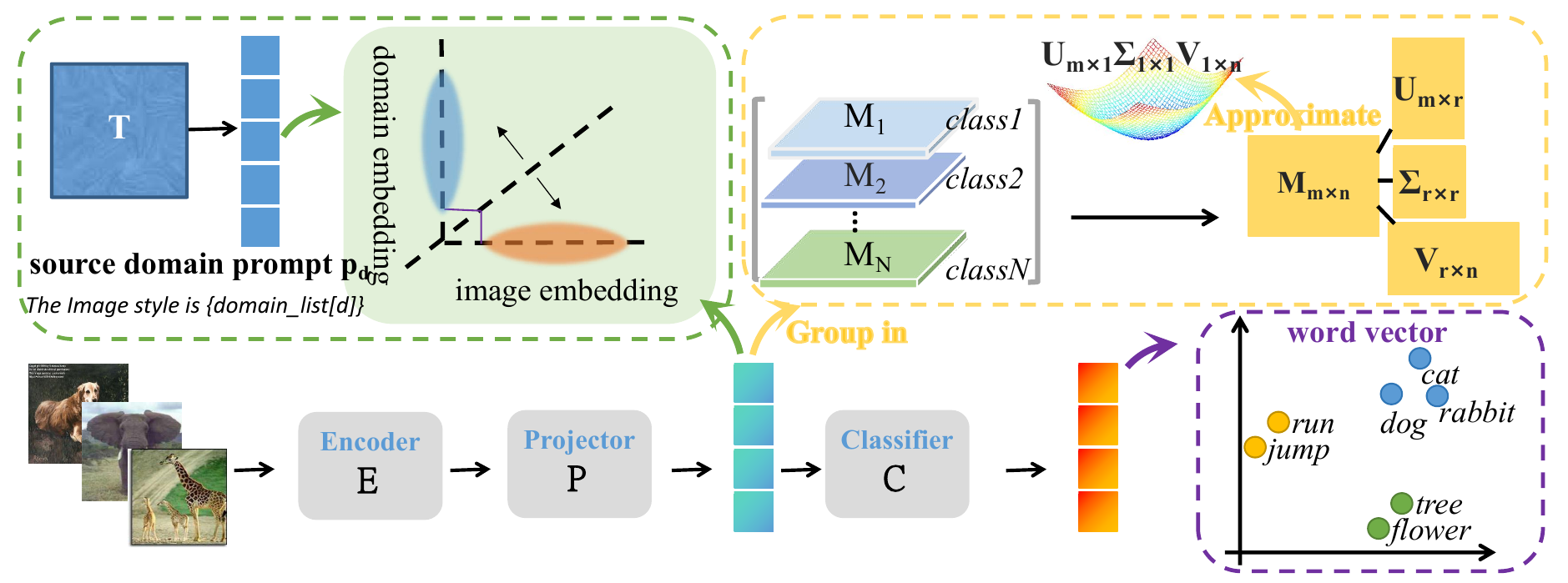}
	\caption{The framework of our proposed VLCA. The method consists of three modules, namely the domain information orthogonal decoupling module based on prompt (green dashed box), the interclass relationship constraint module based on word vectors (purple dashed box), and the intraclass feature consistency module based on low-rank space approximation (yellow dashed box).}
	\label{framewerk}
	% \vspace{-3mm}
\end{figure*}
We integrate language space and vision space, using the semantic space as a bridge domain, to achieve domain generalization through three modules, which are \textbf{domain information orthogonal decoupling module} in vision-language multimodal space, \textbf{word vector interclass constraint module} in language space, and 
 \textbf{low-rank intraclass approximation module} in vision space.
 
{In multimodal space of vision and language}, the orthogonal decoupling module constrains the image feature to be orthogonal to the domain embedding, so as to suppress the domain information in the image feature.
{In language space}, inspired by the consistency of category semantic information in various domains, we design a module based on category relationships. Specifically, 
%the semantic space supervision consists of two main modules, the orthogonal decoupling module and the word embedding interclass constraint module. The orthogonal decoupling module decouple the domain-related information in the image feature and weaken its influence. 
the word vector interclass constraint module provides interclass supervision by mining the semantic relations of category word vectors. {In vision space}, since sample features of the same category are similar and the matrix they form should be low-rank, we aggregate sample features of the same category together to form feature matrices, perform low-rank matrix decomposition, and then make the feature matrix approximate the subspace spanned by the largest singular value.  %project them into the subspace composed of biggest singular vectors.
The overall framework is illustrated in Fig. \ref{framewerk}. In the following section, we first briefly introduce domain generalization, and then introduce these three key components.
\subsection{Problem Definition}
The domain generalization task is defined as follows: given $M$ source domains $S_{train} =\{S_i|i=1,\cdots,M\}$, data of the $i$-th domain are represented as
$S_i =\{x_j^i,y_j^i\}_{j=1}^{n_i}$, where $x_j$ is the image sample and $y_j$ is the category  label. The distribution of data varies between different source domains, $P_{XY}^{i}\neq P_{XY}^j, 1 \leq i \neq j \leq M$. Domain generalization seeks to learn a strong generalizable function $h: \mathcal{X} \rightarrow \mathcal{Y}$ from these M source domains to minimize the error on the unseen test set, where the data distribution of target domain in test set is different from that of source domains. 

\subsection{Feature Decoupling Separates Domain Information}
\begin{figure}[!h]
    \centering
    \includegraphics[width =0.4\textwidth]{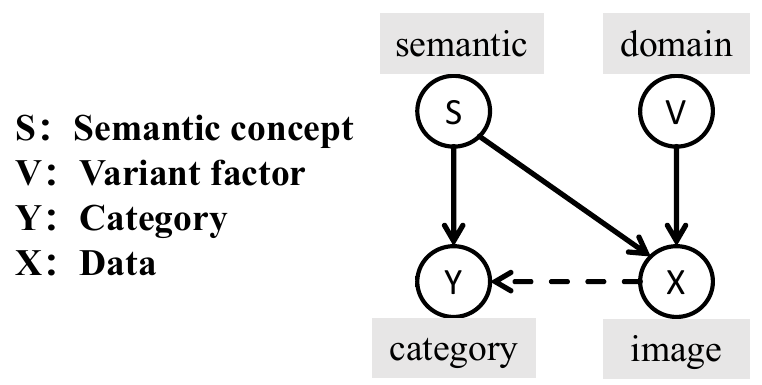}
    \caption{Causal analysis of an image. An image consists of  semantic information and domain information. domain information is a variant factor. The semantic information determines the category.}
    \label{casual}
\end{figure}
From causal analysis, as shown in Fig. \ref{casual}, An image consists of two kinds of information. One is the semantic information, and the other is domain information. But the latter one is a variant factor. Only the semantic information determines the category.
%the class-related semantic information is consistent in each domain, which is conducive to improving the robustness of the model.
\begin{figure}[!h]
    \centering
    \includegraphics[width =0.4\textwidth]{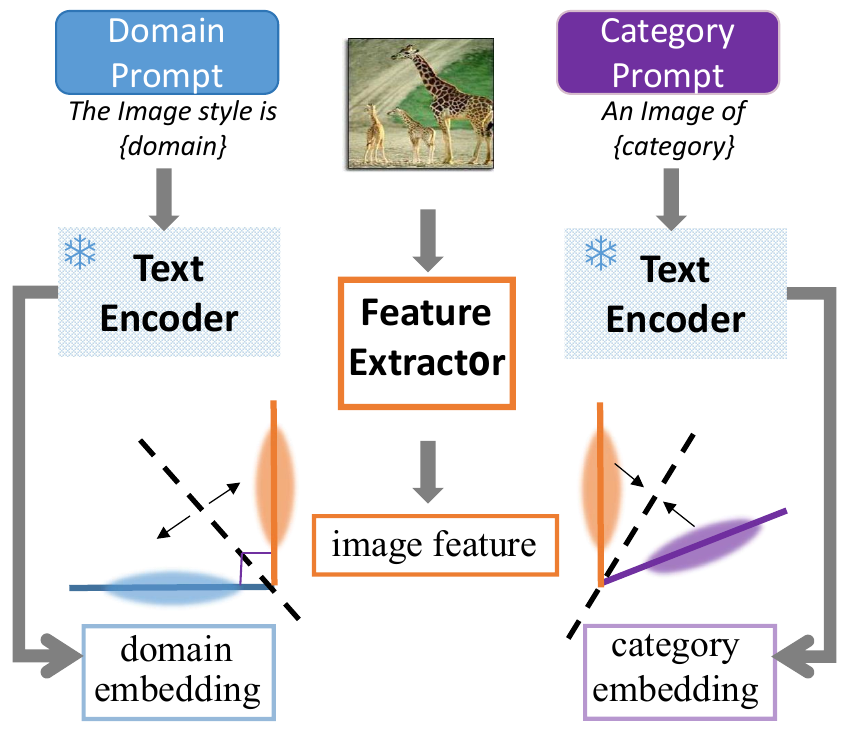}
    \caption{Feature decoupling separates domain information. The domain prompt and category prompt are fed into the feature encoder, yielding the domain embedding and category embedding, respectively. The image feature is constrained to be orthogonal to the domain embedding and aligned with the category embedding.}
    \label{orth}
\end{figure}

Therefore, for any image $I$ belonging to domain $S_i$, we attempt to divide its features into two parts, namely domain-related style information and class-related semantic information.  However, the image features are coupled to each other, and it is difficult to directly realize the decoupling. To address the difficulty, we relax the problem, transforming it into a projection of image feature $\mathbf{F}$ in the directions of domain-related style information $\mathbf{E}_{sty}$ and category-related semantic information $\mathbf{E}_{sem}$, as shown in Fig. \ref{orth}.

The image feature $\mathbf{F}$ is orthogonal to domain-related style information $\mathbf{E}_{sty}$ as formula \ref{decop}. 
\begin{equation}\label{decop}
	\text{proj}_{\mathbf{E}_{sty}}(\mathbf{F}) = \frac{\langle \mathbf{F}, \mathbf{E}_{sty} \rangle}{\langle \mathbf{E}_{sty}, \mathbf{E}_{sty}\rangle} \mathbf{E}_{sty}
\end{equation}
$\mathbf{E}_{sty}$ is got by text encoder of CLIP with prompt \textit{``The Image style is \{domain\_list[d]\}"}. We expect image features $\mathbf{F}$ and the domain-related style information embedding $\mathbf{E}_{sty}$ to be orthogonal, which means the inner product of the domain-related style information embedding $\mathbf{E}_{sty}$ and the image feature $\mathbf{F}$ is zero, as shown in formula \ref{innerproduct}. 
\begin{equation}\label{innerproduct}
	\langle \mathbf{F}, \mathbf{E}_{sty} \rangle =0
\end{equation}
This ensures that the extracted image features do not retain domain-specific information that might influence the generalization ability of the model. 
\begin{figure*}[t]
    \centering
    \includegraphics[width =0.9\textwidth]{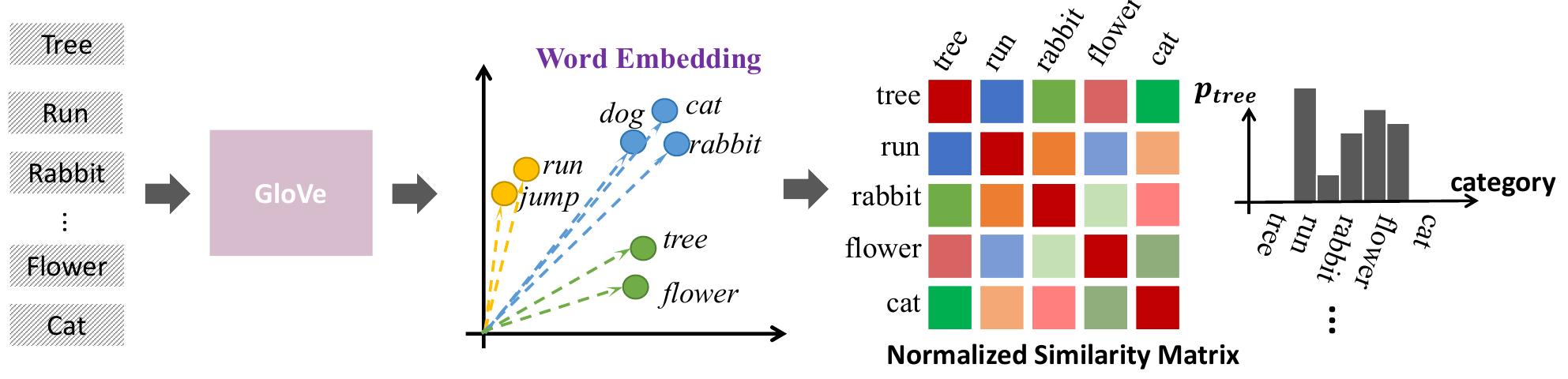}
    \caption{Word vectors construct inter-class relationship. For each category vocabulary, word vectors are obtained through GloVe. Then, the similarity matrix is calculated based on distance of these word vectors, and probability distributions are derived separately for each category.}
    \label{word_embed}
\end{figure*}
Simultaneously, to make the image feature $\mathbf{F}$ and class-related semantic information embedding $\mathbf{E}_{sem}$ as close as possible, we apply semantic consistency supervision to constrain the angle $\theta$ of image feature $\mathbf{F}$ and class-related semantic information embedding $\mathbf{E}_{sem}$ being zero, which means constraining the cosine of the angle $\theta$ between these two vectors to be 1, as shown in formula \ref{cossin}.
\begin{equation}\label{cossin}
	\cos(\theta) = \frac{ \mathbf{F} \cdot \mathbf{E}_{sem}  }{\|\mathbf{F}\| \|\mathbf{E}_{sem}\|}=1
\end{equation}
where $\|{\mathbf{F}}\|$ and $\|{\mathbf{E}_{sem}}\|$ are the norms (lengths) of ${\mathbf{F}}$ and ${\mathbf{E}_{sem}}$ respectively. $\mathbf{E}_{sem}$ is got by text encoder of CLIP with prompt \textit{``An image of  \{category\}"}.

The feature decoupling loss is expressed as two parts, namely the consistent class-related loss and the orthogonal domain-related loss. The loss function is shown below.
\begin{equation}
	\mathcal{L}_{decouple} =  \mathbf{F}\cdot\mathbf{E}_{sty} +(1-\frac{ \mathbf{F} \cdot \mathbf{E}_{sem}  }{\|\mathbf{F}\| \|\mathbf{E}_{sem}\|})
\end{equation}
\subsection{Word Vector Inter-class Constraint}
To further enhance the semantic supervision between categories, word vectors are used to construct semantic distributions among categories. Specifically, with the help of the pre-trained word to vector model GloVe (Global Vectors for Word Representation) \cite{pennington2014glove}, as shown in Fig. \ref{word_embed}, we take each category in category set ${C}$ as the target category ${y_k}$ and calculate the cosine distance ${d_{kl}}$ between the target category ${y_k}$ and other categories ${y_l}$, as in formula \ref{class_sem}. For example, ${y_k}$ is the word vector of \textit{``dog"}, ${y_l}$ is the word vector of \textit{``cat"}, and the ${d_{kl}}$ is the distance between \textit{``dog"} and \textit{``cat"}.
\begin{equation}\label{class_sem}
{d_{kl}} = \frac{{\mathbf{y}_k \cdot \mathbf{y}_l}}{{\|\mathbf{y}_k\| \|\mathbf{y}_l\|}}, \quad \mathbf{y}_k,\mathbf{y}_l \in {C}, 
\end{equation}
Next, the semantic distribution of the target category $y_k$ is constructed by normalizing the set of distances obtained by the target category and other categories and constraining their sum to 1, as shown in formula \ref{dist_sem}.
\begin{equation} \label{dist_sem}
	{p_l} = \frac{{d_{kl}}}{\sum_{{j}=1}^{{c}} {d_{kj}}},
\end{equation}
where $c$ is the category number of the  category set ${C}$.
\\
Then, the semantic probability distribution of the target category $y_k$ is represented as $\mathbf{P}_k = \{p_1,\cdots, p_c\}$. Similarly, the semantic probability distributions for other categories are calculated. Use these distributions as supervision to calculate the semantic relationship loss $\mathcal{L}_{semantic}$ between the semantic probability distribution $\mathbf{P}$ and the network output $\mathbf{f}$ shown in below.
\begin{equation}
	{\mathcal{L}_{semantic}} =\sum_{k=1}^{m} \mathbf{P}_{k} \log \left( \frac{\mathbf{P}_k}{\mathbf{f}_k} \right)
\end{equation}
where $m$ means that there are m samples in a batch.
\subsection{Low-rank Decomposition Intra-class Constraint}
%Images of the same class from different domains vary widely in image space, nevertheless have a low rank in feature space due to feature similarity. 
\begin{figure}[!h]
    \centering
    \includegraphics[width =0.5\textwidth]{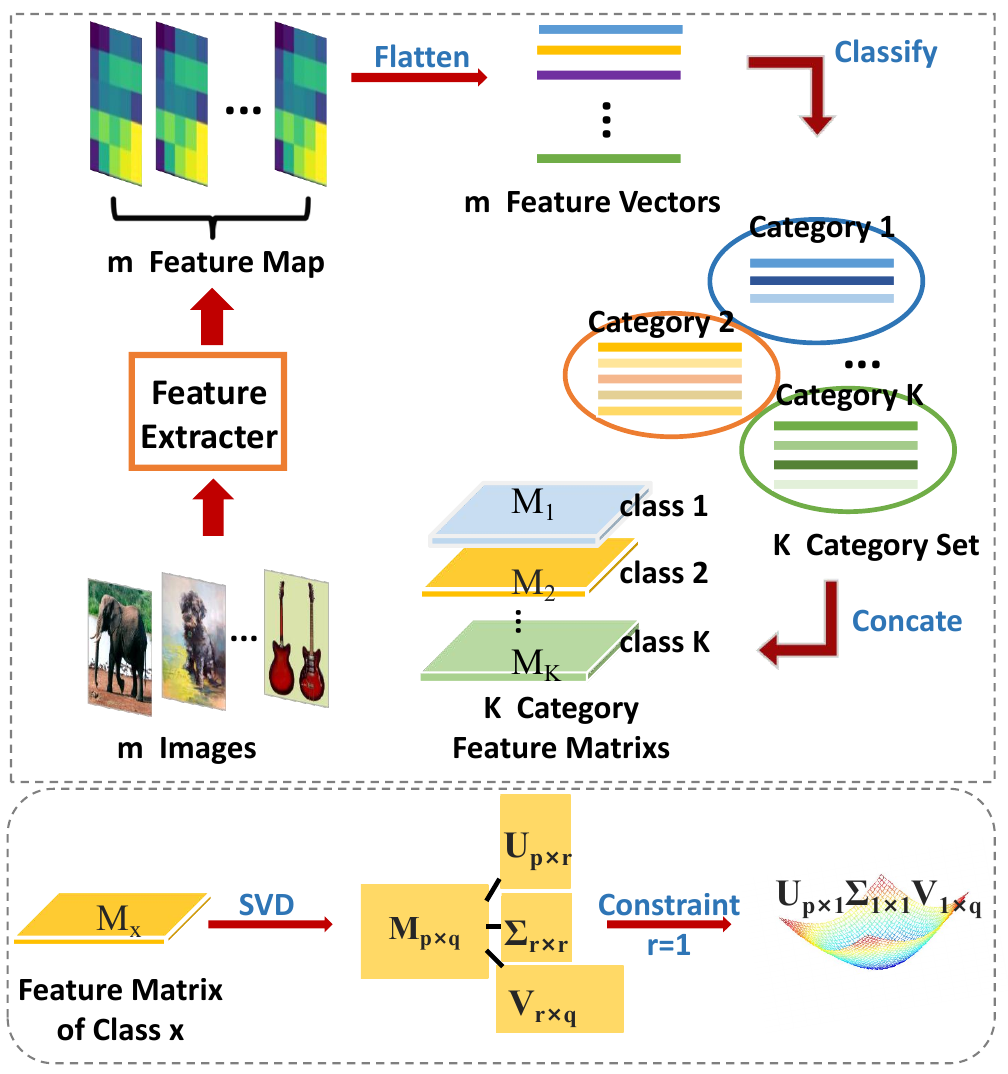}
    \caption{Low-rank decomposition realizes intra-class features constraint. Given $m$ images as input, $m$ feature maps are extracted by feature extractor. Then $m$ feature maps are flattened as $m$ feature vectors. These feature vectors are categorized into specific classes. For each class, we concate its vectors to form a matrix $M$. For any given matrix $M_x$, Singular Value Decomposition (SVD) is applied to obtain left singular vectors matrix $U$, right singular vectors matrix $V^T$ and singular values matrix $\Sigma$. By constraining $r=1$, enforces the inter-row correlation of the matrix $M_x$, thus constraining the consistency of features across different domain images of the same category to approximate the ideal feature space spanned by the largest singular value.  }
    \label{fig:enter-label}
\end{figure}
Although images of the same class exhibit significant differences across domains in image space, they maintain low rankness in feature space due to their feature similarity.
To enhance the feature consistency within categories, we form a matrix using the image features with the same category label in one batch. 

Let \( \mathcal{F} \) be the feature matrix of a batch, where $ \mathcal{F} = [\mathbf{F}_1^{l_{x_1}}, \mathbf{F}_2^{l_{x_2}}, \cdots, \mathbf{F}_m^{l_{x_m}}]^T$. $l_x$ is the category label of the feature. We perform row permutations on the feature matrix \( \mathcal{F} \), grouping rows with the same category together, as shown in the following equation
\begin{equation}
	\mathcal{F} =
	\begin{bmatrix}
		\mathbf{F}_1^{l_{x_1}} \\
		\mathbf{F}_2^{l_{x_2}}\\
		\vdots \\
		\mathbf{F}_m^{l_{x_m}}
	\end{bmatrix}
=\begin{bmatrix}
	\mathbf{F}_p^{l_{i}} \\
	\vdots\\
	\mathbf{F}_q^{l_{j}}\\
	\vdots \\
	\mathbf{F}_r^{l_{k}}
\end{bmatrix}
\equiv \begin{bmatrix}
	\mathbf{M}_{i} \\
	\mathbf{M}_{j}\\
	\vdots \\
	\mathbf{M}_{k}
\end{bmatrix}
\end{equation}
where $\mathbf{M}_x$ represents the matrix composed of sample features of the category $x$.\\
For any given category feature matrix \(M\), applying Singular Value Decomposition (SVD) results in the factorization:
\begin{equation}
\mathbf{M} = \mathbf{U} \mathbf{\Sigma}\mathbf{ V}^T = \mathbf{\sigma_1u_1v_1}+\mathbf{\sigma_2u_2v_2}+\cdots+\mathbf{\sigma_ru_rv_r}
\end{equation}
where \(\mathbf{U}\) is the left singular matrix, \(\mathbf{\Sigma}\) is the diagonal matrix of singular values, and \(\mathbf{V}\) is the right singular matrix.\\
Since the matrix $\mathbf{M}$ is composed of features from the same category of images, there are common patterns among these features, resulting in shared similar feature vectors. This ultimately reduces the rank $r$ of the matrix. Ideally, the rank of matrix $\mathbf{M}$ should be 1. 
\begin{equation}
	\mathbf{M}_{ideal} =\mathbf{\sigma_1u_1v_1} = \mathbf{U}_{m\times1} \mathbf{\Sigma}_{1\times1} \mathbf{V}^T_{1\times n}
\end{equation}
In other words, the first principal component $\mathbf{\Sigma}(1)$ carries the majority of the information in the matrix. Therefore, we constrain the matrix $\mathbf{M}$ to approach the ideal matrix $\mathbf{M}_{ideal}$ by imposing a constraint on the rank $r$ of the matrix, aiming for it to approach 1.
By constraining $r=1$, achieving $rank(M_x)=1$ further enforces the inter-row correlation of the class matrix, thus constraining the consistency of features across different domain images.

\begin{equation}
		{\mathcal{L}_{approximate}} = {r}-1
\end{equation}
%which represents a common pattern between data samples.     
%which represents a common pattern between features.
%Project the category feature matrix \(M\) onto the common feature space spanned by the left singular vectors \(U\),
%\begin{equation}
%	M_{proj} = UU^T M
%\end{equation}
%where $M_{proj}$ represents the matrix obtained after projection, which reflects the representation of category features in the common feature space.
%Then, we construct the intraclass feature consistency loss,
%\begin{equation}
%	L_{project} =  \sum_{i=1}^{m}\sum_{j=1}^{n}(M_{ij} - (UU^TM)_{ij})^2
%\end{equation}
Combining all the loss functions together, we define full objective as:
\begin{equation}
	{\mathcal{L}} = {\mathcal{L}}_{cls}+ \alpha \cdot ({\mathcal{L}_{decouple}}+{\mathcal{L}_{semantic}})+\beta \cdot   {\mathcal{L}_{approximate}}
\end{equation}
where $\alpha$ and $\beta$ control the weights of the decouple loss and the semantic loss.
\section{Experiments}
We demonstrate the superiority of the proposed method on multiple DG benchmarks in this section. In addition, ablation experiments are performed on vision supervision and language constraints, respectively.
\subsection{Datasets and Setting}
The proposed method is evaluated on PACS \cite{li2017deeper}, Office-Home \cite{venkateswara2017deep} VLCS \cite{fang2013unbiased} and  Terraincognita\cite{beery2018recognition} DG benchmarks. We use the train-validation split following \cite{zhou2021domain, zhou2022domain}. The detailed experimental parameters are set as follows. We train 50 epochs using SGD, batch size of 16, and weight decay of 5e-4. The initial learning rate is 0.001, decreasing by 0.1 every 40 epochs. The parameter settings are the same as \cite{carlucci2019domain,zhou2020deep,xu2021fourier}. We use the image augmentation as \cite{xu2021fourier}. Domain prompt embedding is generated by the text encoder of contrastive language-image pretraining (CLIP) \cite{radford2021learning}. For simplicity, we chose the \textit{``RN101"} CLIP pre-trained model\footnote{\url{https://github.com/openai/CLIP}} as the text encoder. For other CLIP pre-trained text models, we show their experimental results in  Section \ref{abla_clip}. The ResNet-18 and ResNet-50 pretrained on ImageNet-1k dataset are used as the backbone following previous methods. In the PACS and Office Home datasets, the prompt is \textit{``The image style is \{domain\}"} and in the VLCS dataset, the prompt is \textit{``The image style is from dataset \{domain\}"}. Category word vectors are obtained through pre-trained GloVe \cite{pennington2014glove} models. We chose the simplest GloVe model, \textit{``glove.6B.50d"}\footnote{\url{https://nlp.stanford.edu/projects/glove}}. The performance of other pre-trained word embedding models is provided in Section \ref{abla_wordvec}. Since the glove word vector model can only express a single lowercase word, it cannot express compound words, such as ``alarm clock" in the OfficeHome dataset. Through the utilization of algebraic operations in the text embedding space (e.g. king-man+woman  approaches the word representation of queen), we use the addition of two parts to express the semantic embedding of compound words. For words that are not compound words but are not retrieved, we take the approach of synonym substitution, e.g. football $\Rightarrow$ soccer, flipflop $\Rightarrow$ slipper.
For all experiments, the $\alpha$ and $\beta$ are set to 0.2. 
\subsection{Comparison with existing DG methods}
% To verify the effectiveness of the proposed method, we compare it with the following domain generalization methods.
\subsubsection{Multi-source Domain Generalization on PACS}
\begin{table*}[!h]
	\caption{Leave-one-domain result on PACS with ResNet-18. The best results are bolded. The second-best results are underlined.}
	\centering
	%\resizebox{0.485\textwidth}{!}{%
		\begin{tabular}{@{}l|c|cccc|c@{}}
			\toprule
			{ Methods} &Venue& { Art} & { Cartoon} & { Photo} & { Sketch} & { Avg.} \\ \midrule
			MetaReg\cite{balaji2018metareg} &2018 NeurIPS&83.7&77.2&95.5 &70.3&81.7\\
			JiGen\cite{carlucci2019domain}&2019 CVPR&79.4 &75.3&96.0  &71.4 &80.2\\
			EISNet\cite{wang2020learning}&2020 ECCV&81.9 &76.4&95.9  &74.3 &82.2\\
			RSC\cite{huang2020self} &2020 ECCV &83.4 &80.3 &96.0 &\textbf{80.9} &82.2\\
			FACT\cite{xu2021fourier} &2021 CVPR&\underline{85.4} &78.4 &95.2 &79.2 &\underline{84.5}\\
			Pro-RandConv\cite{choi2023progressive}&2023 CVPR&83.2 &\textbf{81.1} &{96.2} &76.7 &84.3\\
                JVINet\cite{chu2024joint}&2024 PR&81.2&79.0&\textbf{96.6}&74.9&82.9\\
			%VNE\cite{kim2023vne}&2023CVPR &88.6 &79.9 &96.7 &82.3 &86.9\\
			{Ours} & -&\textbf{85.79}&\underline{80.59}&\underline{96.29}&\underline{80.76}&\textbf{85.86}\\
			\bottomrule
	\end{tabular}
 %}
\label{PACS18}
\end{table*}
\begin{table*}[!h]
	\caption{Leave-one-domain result on PACS with ResNet-50. The best results are bolded. The second-best results are underlined.}
	\centering
	%\resizebox{0.485\textwidth}{!}{%
	\begin{tabular}{@{}l|c|cccc|c@{}}
		\toprule
		{ Methods} &Venue& { Art} & {  Cartoon} & {  Photo} & {  Sketch} & {  Avg.} \\ \midrule
		MetaReg\cite{balaji2018metareg} &2018 NeurIPS  & 87.2   		& 79.2  &97.6 & 70.3 & 83.6 \\
		MASF\cite{dou2019domain} &2019 NeurIPS		   & 82.9   		& 80.5 & 95.0  & 72.3 & 82.7 \\
		EISNet\cite{wang2020learning}&2020 ECCV  	   & 86.7  			& 81.5 & 97.1 & 78.1  & 85.8   \\
		RSC\cite{huang2020self}    &2020 ECCV   	   & 87.9 			& 82.2& 97.9  & 83.4   & 87.8   \\
		FACT\cite{xu2021fourier} &2021 CVPR  		   & 89.6	  		& 81.8 &96.7      &{84.5}	&88.2		\\
		PCL\cite{yao2022pcl}   &2022 CVPR    	    & {90.2}  & 83.9  & \underline{98.1}    & 82.6  & {88.7}  \\
		POEM\cite{jo2023poem} &2023 AAAI 			   & 89.4 			& 83.0 &\textbf{98.2} &83.2&88.5\\
	Pro-RandConv\cite{choi2023progressive}&2023 CVPR   & 89.3			& {84.1} &97.8&81.9 &88.3\\
	VNE\cite{kim2023vne}&2023 CVPR&90.1 &83.8 &97.5 &81.8 &88.3\\
 CBDMoE\cite{xu2024cbdmoe}  & 2024 TMM  & 87.3 & 85.0 & \textbf{98.2}  & \underline{85.3} & 88.9 \\
NormAUG\cite{qi2024normaug} & 2024 TIP  & 88.9 & \textbf{86.0}    & 97.2 & \textbf{86.0} & \underline{89.5} \\ 
HDA-LMS\cite{ng2024improving}&2024 NN &87.7&81.2&97.8&81.4&87.0\\
DBAM\cite{li2024takes}&2024 NN&\underline{91.0}&84.0&\underline{98.1}&83.2&89.1\\
		Ours&- 										   & \textbf{91.94}	&\underline{84.30}	&97.84	&{84.83}	&\textbf{89.73}\\
		\bottomrule
	\end{tabular}
%}
	\label{PACS50}
\end{table*}
The results are shown in TABLE \ref{PACS18} and TABLE \ref{PACS50}. Unlike other comparison methods, our proposed method achieves optimal or suboptimal results in all domains. This is because we use dual supervision of language and vision features, which can be well constrained in the ``Photo" domain and ``Art" domain with rich texture and details, while the semantic information plays a good role in alignment for the more abstract and less detailed ``Cartoon" domain and ``Sketch" domain.
\subsubsection{Multi-source Domain Generalization on OfficeHome}
\begin{table*}[!h]
		\caption{Leave-one-domain result on Office-Home with ResNet-18. The best results are bolded. The second-best results are underlined.}
	\centering
	%\resizebox{0.485\textwidth}{!}{
		\begin{tabular}{@{}l|c|cccc|c@{}}
		\toprule
		{  Method} & Venue& {  Art} & {  Clipart} & {  Product} & {  Real} & {  Avg.} \\ \midrule
		CCSA\cite{motiian2017unified}&2017 ICCV &59.9 &49.9 &74.1 &75.7 &64.7\\
		MMD-AAE\cite{li2018domain}&2018 CVPR&56.5 &47.3 &72.1 &74.8 &62.7\\
		JiGen\cite{carlucci2019domain}&2019 CVPR&53.0 &47.5 &71.5 &72.8 &61.2\\
		DDAIG\cite{zhou2020deep}&2020 AAAI&59.2 &\underline{52.3} &\textbf{74.6} &\underline{76.0} &65.5\\
		L2A-OT\cite{zhou2020learning}&2020 ECCV &\textbf{60.6} &50.1 &74.8 &\textbf{77.0} &65.6\\
		RSC\cite{huang2020self}    &2020 ECCV &58.4 &47.9 &71.6 &74.5 &63.1\\
		% \textcolor{blue}{FACT\cite{xu2021fourier}} &2021 CVPR&60.3 &\underline{54.9} &\underline{74.5} &\underline{76.6} &\textbf{66.6}\\
  %       \textcolor{blue}{CIRL\cite{lv2022causality}}&2022 CVPR&61.48 &55.28 &75.06 &76.64 &67.12\\
		%PCL\cite{yao2022pcl}&2022 CVPR&62.10 &58.22 &77.38 &77.98 &68.92\\
		VNE\cite{kim2023vne}&2023 CVPR& \underline{60.4} &54.7 &73.7 &74.7 &\underline{65.9}\\
            HDA-LMS\cite{ng2024improving}&2024 NN &58.5& 50.8& 73.4& 74.5& 64.3\\
		Ours &- &59.33&\textbf{56.45} &\underline{74.16}&74.94 &\textbf{66.2}\\
		\bottomrule 
		\end{tabular}
	%}
\label{OfficeHome18}
\end{table*}
\begin{table*}[!h]
	\caption{Leave-one-domain result on Office-Home with ResNet-50. The best results are bolded. The second-best results
are underlined.}
	\centering
	\resizebox{0.99\textwidth}{!}{
	\begin{tabular}{@{}l|c|cccc|c@{}}
		\toprule
		{  Method} & Venue& {  Art} & {  Clipart} & {  Product} & {  Real} & {  Avg.} \\ \midrule
		RSC\cite{huang2020self}& 2020 ECCV  & 60.7  & 51.4    & 74.8   &75.1  &65.5\\
		MMD\cite{li2018domain} & 2018 CVPR  & 60.4  & 53.3    & 74.3   &77.4  &66.4\\
		MTL\cite{blanchard2021domain} & 2021 JMLR   & 61.5   &52.4   & 74.9 &76.8   & 66.4  \\
		VREx\cite{krueger2021out}    &2021 ICML  & 60.7  & 53.0  & 75.3  & 76.6  & 66.4                        \\
		MLDG\cite{li2018learning}  &2018 AAAI  & 61.5   & 53.2  & 75.0  & 77.5    & 66.8                        \\
		I-Mixup\cite{xu2020adversarial}   &2020 AAAI & 62.4 & 54.8   & 76.9  & 78.3  & 68.1                        \\
		SagNet\cite{nam2021reducing}  &2021 CVPR & 63.4  & 54.8   & 75.8 & 78.3 & 68.1                        \\
		SWAD\cite{cha2021swad}   &2021 NeurIPS   & \underline{66.1} & 57.7  & 78.4  & \underline{80.2}  & 70.6 \\
		%PCL\cite{yao2022pcl}&2022 CVPR	&67.3&59.9&78.7&\textbf{80.7}&71.6\\
		POEM\cite{jo2023poem} & 2023 AAAI&64.1&53.9&76.2&77.6&68.0\\
		VNE\cite{kim2023vne}&2023 CVPR &\textbf{66.6} &\underline{58.6} &\textbf{78.9} &\textbf{80.5} &\underline{71.1}\\
            JVINet\cite{chu2024joint}&2024 PR&-&-&-&-&68.3\\
		Ours       &-                   & 65.43 & \textbf{61.37}   & \underline{78.55}   & 79.25 & \textbf{71.15}              \\ \bottomrule
	\end{tabular}
}
	\label{OfficeHomeresnet50}
\end{table*}
The results are shown in TABLE \ref{OfficeHome18} and TABLE \ref{OfficeHomeresnet50}. The experimental results show that our method achieves competitive results, but the improvement is not obvious enough. The reason may be that, firstly, the OfficeHome dataset has 65 classes, and in order to fairly compare with other methods, we set the batch size to 16, which affects the constraints of the intraclass feature consistency module on the network. Secondly, there are many compound word labels in the dataset, and these words are not retrieved in the word vector model, which affects the network performance. As shown in Fig. \ref{batchsize} of Section \ref{abla_batchsize}, the experimental performance improves with the increase of the batch size.
% \usepackage{booktabs}
\subsubsection{Multi-source Domain Generalization on VLCS}
\begin{table*}[!h]
	\caption{Leave-one-domain result on VLCS with ResNet-18 and ResNet-50. The best results are bolded.}
	\centering
	% \resizebox{0.8\textwidth}{!}{
	\begin{tabular}{@{}ccccccc@{}}
		\toprule
		Method                         & Venue     & C     & L     & S     & V                          & Avg.  \\ \midrule
		\multicolumn{7}{c}{ResNet 18}                                                                           \\ \midrule
		\multicolumn{1}{l|}{DeepAll\cite{zhou2020deep}}   & 2020 AAAI & 91.9  & 61.8  & 68.8  & \multicolumn{1}{c|}{67.5}  & 72.5  \\
		\multicolumn{1}{l|}{RSC\cite{huang2020self}}       & 2020 ECCV & 95.8  & 63.7  & 72.1  & \multicolumn{1}{c|}{71.9}  & 75.9  \\
		\multicolumn{1}{l|}{MMLD\cite{matsuura2020domain}}      & 2020 AAAI & 97.0  & 62.2  & \underline{72.5}  & \multicolumn{1}{c|}{73.0}  & 76.2  \\
		\multicolumn{1}{l|}{StableNet\cite{zhang2021deep}} & 2021 CVPR & 96.7  & 65.4  & \textbf{75.0}  & \multicolumn{1}{c|}{73.6}  & 77.7  \\
		\multicolumn{1}{l|}{MVDG\cite{zhang2022mvdg}}      & 2022 ECCV & \textbf{98.4}  & 63.8  & 71.1  & \multicolumn{1}{c|}{75.3}  & 77.2  \\
		\multicolumn{1}{l|}{VNE\cite{kim2023vne}}       & 2023 CVPR  & 97.5  & \underline{65.9}  & 70.4  & \multicolumn{1}{c|}{\textbf{78.4}}  & \underline{78.1}  \\
		\multicolumn{1}{l|}{Ours}      &-           & \underline{97.88} & \textbf{68.30} & 72.46 & \multicolumn{1}{c|}{\underline{75.89}} & \textbf{78.63} \\ \midrule
		\multicolumn{7}{c}{ResNet 50}                                                                           \\ \midrule
		\multicolumn{1}{l|}{VNE\cite{kim2023vne}}       & 2023 CVPR  & \textbf{99.2}  & 63.7  & 74.4  & \multicolumn{1}{c|}{\textbf{81.6}}  & 79.7  \\
            \multicolumn{1}{l|}{JVINet\cite{chu2024joint}}&2024 PR&-&-&-&-&79.1\\
		\multicolumn{1}{l|}{Ours}      &-           & 98.80 & \textbf{67.78} & \textbf{77.12} & 78.41                      &  \multicolumn{1}{|c}{\textbf{80.53}} \\ \bottomrule
	\end{tabular}
  %}
\label{VLCS}
\end{table*}
The results are shown in TABLE \ref{VLCS}. In the LabelMe dataset domain, the experimental results on ResNet 18 and ResNet 50 networks are significantly improved, with an average increase of three points. The VLCS dataset only includes 5 categories, so that more common patterns of samples of the same category can be found in low-rank approximation, so as to effectively suppress the interference factors and specific information in the image features.
%In the VOC2007 dataset domain, the general reason for the experimental results is that there are many interferences in the images and the classification target is not clear.
\subsubsection{Multi-source domain generalization on Terraincognita}
\begin{table*}[!h]
\centering
\caption{{Leave-one-domain result on Terraincognita with ResNet-50. The best results are bolded.}}
\resizebox{0.96\textwidth}{!}{
\begin{tabular}{@{}l|ccccc|c@{}}
\toprule
Method      & Venue        & Location100 & Location38 & Location43 & Location46 & Avg  \\ \midrule
RSC\cite{huang2020self}         & 2020 ECCV    & 50.2        & 39.2       & 56.3       & \underline{40.8}       & 46.6 \\
I-Mixup\cite{xu2020adversarial}     & 2020 AAAI    & \textbf{59.6}        & 42.2       & 55.9       & 33.9       & 47.9 \\
VREx\cite{krueger2021out}        & 2021 ICML    & 48.2        & 41.7       & 56.8       & 38.7       & 46.4 \\
SagNet\cite{nam2021reducing}      & 2021 CVPR    & 53.0        & 43.0       & 57.9       & 40.4       & 48.6 \\
SWAD\cite{chattopadhyay2020learning}        & 2021 NeurIPS & 55.4        & 44.9       & \underline{59.7}       & 39.9       & 50.0 \\
%PCL         & 2022 CVPR    & \underline{58.7}        & \underline{46.3}       & \underline{60.0}      & \textbf{43.6}      & \textbf{52.1} \\
Ratatouille\cite{rame2023model} & 2023 CVPR    & \underline{57.9}        & \textbf{50.6}       & \textbf{60.2}       & 39.2       & \textbf{52.0} \\
JVINet\cite{chu2024joint}&2024 PR&-&-&-&-&46.8\\
Ours        & -             & 59.61      & \underline{44.96}     & 57.08      & \textbf{42.14}      & \underline{50.94}     \\ \bottomrule
\end{tabular}}
\label{terrain}
\end{table*}
The results are shown in TABLE \ref{terrain}. Our approach demonstrates competitive performance on long-tailed distribution datasets. This is attributed to the integration of semantic supervision via word embeddings and the text encoder, which mitigates the impact of imbalanced sample quantities. However, limitations in sample counts for certain classes hinder the effectiveness of our low-rank approximation module. 
\subsubsection{Single Domain Generalization Results}
%The single domain generalization results on PACS dataset with ResNet 18 are shown in Table \ref{single}. The proposed algorithm has a good performance in the transfer between more abstract image domains, such as in the A→S and C→S. Since our method is semantic-based,  the presence of excessive textures and intricate details in the photo domain images hinders the extraction of high-level semantic features. Therefore, the performance of the method in photo domain transfer is quite ordinary.
The results of single domain generalization on the PACS dataset using the ResNet 18 model are presented in TABLE \ref{single}. Notably, our proposed algorithm exhibits commendable performance when transferring knowledge across inherently more abstract image domains, as exemplified in the A→S and C→S scenarios. The noteworthy performance is attributed to the semantic result-based nature of our method. Nevertheless, it is imperative to acknowledge that the abundance of intricate textures and fine-grained details within images from the photo domain poses a challenge in the extraction of high-level semantic features. Consequently, the performance of our method in the domain transfer involving photo images is observed to be comparatively standard.
\begin{table*}[!h]
	\caption{Single-source Domain Generalization on PACS with ResNet-18.(A: Art Painting, C: Cartoon, S: Sketch, P: Photo). JiGen  \cite{carlucci2019domain} results are
		reproduced with their official code. ADA \cite{volpi2018generalizing} and SagNet \cite{nam2021reducing} results are reported based on implementations from \cite{nam2021reducing}. }
	%GeoTexAug \cite{liu2022geometric} is reported based on implementations from \cite{liu2022geometric}. 
	\centering
	\resizebox{1\textwidth}{!}{
		\begin{tabular}{c|c|cccccccccccc|c}
			\toprule
			Methods&Venue& A→C& A→S& A→P& C→A &C→S& C→P& S→A &S→C &S→P &P→A &P→C& P→S& Avg. \\
			\midrule 
			%		ResNet-18 & & 56.5 & 45.6 & 95.8 & 59.1  & 62.6 & 84.0 & 20.9 & 37.1 & 27.4 & 60.7 & 25.4 & 30.2 & 50.4       \\
			JiGen \cite{carlucci2019domain}&2019 CVPR & 57.0& 50.0 &96.1& 65.3 &65.9 &85.5 &26.6 &41.1& 42.8& 62.4 &27.2 &35.5 &54.6  \\
			ADA \cite{volpi2018generalizing}&2018 NeurIPS &64.3 &58.5& 94.5 &66.7 &65.6 &83.6 &37.0 &58.6 &41.6 &65.3 &32.7 &35.9 &58.7   \\
			SagNet \cite{nam2021reducing} &2021 CVPR &67.1& 56.8 &95.7 &72.1& 69.2 &85.7 &41.1 &62.9& 46.2& 69.8 &35.1 &40.7 &61.9 \\
			GeoTexAug \cite{liu2022geometric}&2022 CVPR &\textbf{67.4} &51.7 &\textbf{97.1} &\textbf{79.8} &66.4 &\textbf{89.9} &50.6 &\textbf{70.5} &\textbf{58.8} &\textbf{69.4} &\textbf{38.7} &39.1 &\textbf{65.0}  \\
			Ours &-&65.70 &\textbf{65.18}&96.11&77.25&\textbf{70.73}&87.49&\textbf{52.44}&62.54&55.15&66.06&34.98&\textbf{44.95}&64.88\\	\bottomrule  
		\end{tabular}
	}
	\label{single}
\end{table*}
\subsection{Ablation Study and Analysis}
% \usepackage{booktabs}
% \usepackage{booktabs}
\begin{table}[htbp]
    \centering
	\caption{Ablation study. We study the influence of three different components of our method. cls: the classification loss, semantic: word vectors inter-class supervision module, decouple: domain embedding orthogonal decouple module, approximate: feature matrix approximate module. }
	%\resizebox{0.5\textwidth}{!}{
	\begin{tabular}{@{}cccc|c@{}}
		\toprule
		\multicolumn{4}{c|}{Model Component}  & \begin{tabular}[c]{@{}l@{}}Target Domain\\ Accuracy (\%)\end{tabular}                       \\ \midrule
		\multicolumn{1}{c|}{Method}    & $\mathcal{L}_{semantic}$ & $\mathcal{L}_{decouple}$ & $\mathcal{L}_{approximate}$ & Avg.  \\ \midrule
		\multicolumn{1}{c|}{Baseline}            & -             & -              & -        & 80.16 \\ 
		\multicolumn{1}{c|}{Model A}            & \checkmark               & -              & -       & 84.35 \\
		%\multicolumn{1}{c|}{Model B}  &  \checkmark          & -               & \checkmark     & -          & 85.64$\pm$0.41         & 79.01$\pm$0.38   & 96.29$\pm$0.16 & \multicolumn{1}{c|}{80.40$\pm$0.47}  & 85.34 \\
		%\multicolumn{1}{c|}{Model C}  & \checkmark           & -               & -              & \checkmark   & \textbf{86.72$\pm$0.23}         & 79.27 $\pm$0.55  & 96.05$\pm$0.21 & \multicolumn{1}{c|}{80.71$\pm$0.76}  & 85.69 \\
		\multicolumn{1}{c|}{Model D}            & \checkmark       & \checkmark     & -         & 84.91 \\
		%\multicolumn{1}{c|}{Model E}  & \checkmark           &\checkmark      & -              & \checkmark   & 85.60$\pm$0.18         & 79.79$\pm$0.36   & 96.61$\pm$0.15 & \multicolumn{1}{c|}{\textbf{81.12$\pm$0.34}}  & 85.78 \\
		%\multicolumn{1}{c|}{Model F}  & \checkmark           & -               & \checkmark     & \checkmark  & 85.64$\pm$0.20         & 79.61$\pm$0.29   & \textbf{96.65$\pm$0.14} & \multicolumn{1}{c|}{80.07$\pm$0.37}  & 85.24 \\\midrule
		\multicolumn{1}{c|}{Ours}            & \checkmark             & \checkmark      & \checkmark     & \textbf{85.86} \\ \bottomrule
	\end{tabular}
 %}
\label{ablation}
\end{table}
\subsubsection{Impact of different components}
We perform a comprehensive ablation study to examine the contribution of each component within our model, as detailed in TABLE \ref{ablation}. 
Beginning with baseline, model A is with word vector semantic module. Model D is with word vector semantic module and CLIP domain information decouple module. The proposed method is all with word vector semantic module, CLIP domain information decouple moduler and same-category-feature approximation module.
\subsubsection{3D t-SNE Visualization on Category}
\begin{figure*}[!ht]
	\centering
	\subfloat{
		\begin{minipage}[t]{0.3\linewidth}
			\includegraphics[width=1\linewidth]{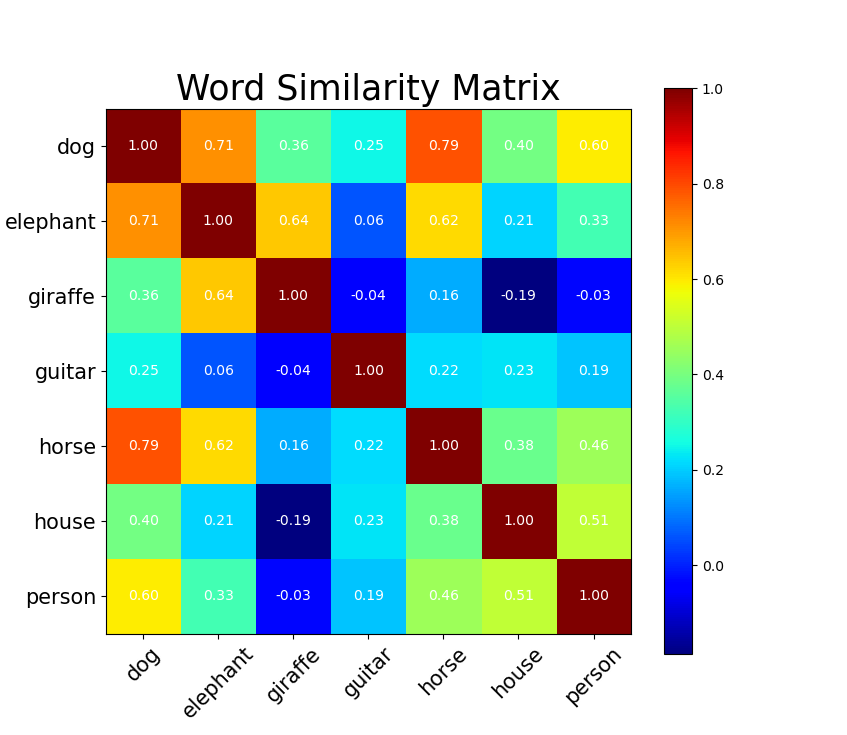}  
		\end{minipage}
		\hspace{-10mm}
		\begin{minipage}[t]{0.3\linewidth}
			\includegraphics[width=1\linewidth]{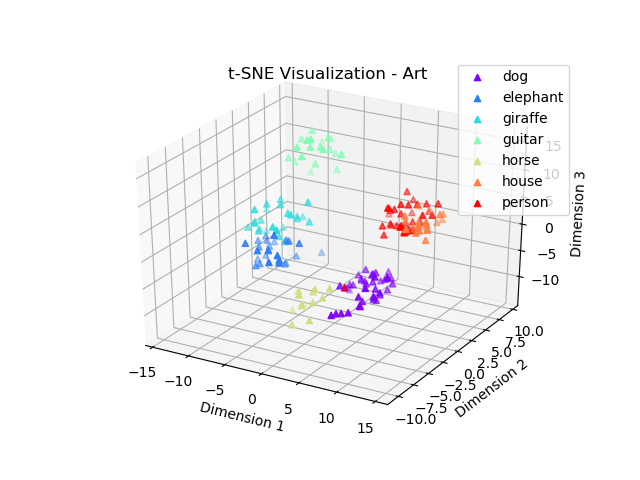}  
		\end{minipage}}
      \vspace{-5mm}      
            \subfloat{
		\begin{minipage}[t]{0.3\linewidth}
			\includegraphics[width=1\linewidth]{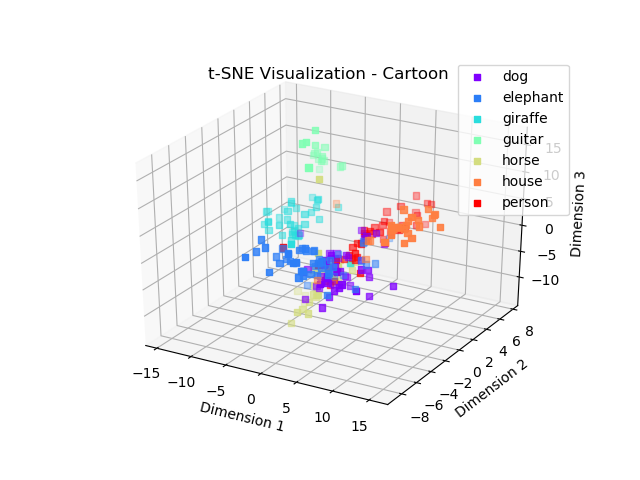}  
	\end{minipage}
       \hspace{-4mm}
		\begin{minipage}[t]{0.3\linewidth}
			\includegraphics[width=1\linewidth]{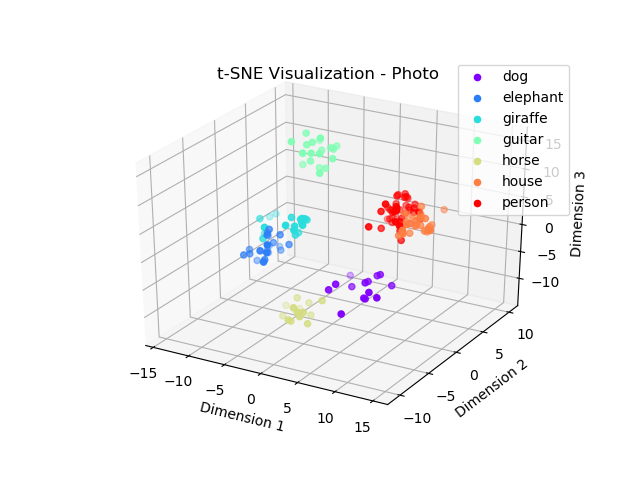}  
		\end{minipage}
		\hspace{-4mm}
		\begin{minipage}[t]{0.3\linewidth}
			\includegraphics[width=1\linewidth]{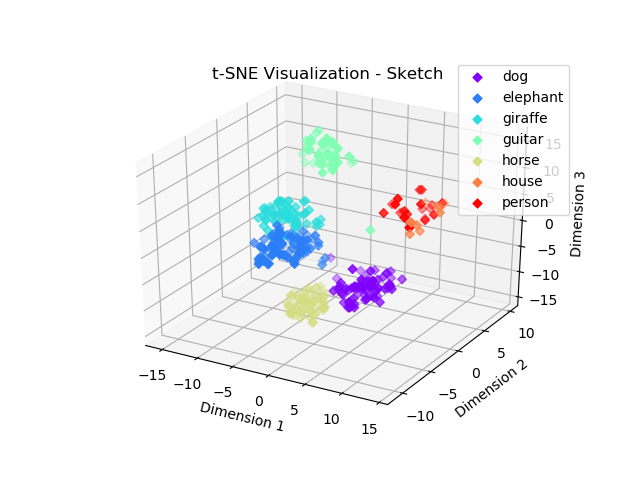}  
	\end{minipage}}
	\caption{The 3D t-SNE visualization of leave-one-domain models on PACS with ResNet 18. The leave-one-domain is set to Art. It can be visually observed that the features of giraffe are close to the features of epephant, as their word vectors are close. }
	\label{3dtsne}
\end{figure*}

The 3D t-SNE visualization of leave-one-domain models in Cartoon domain on PACS with ResNet 18 is shown in Fig. \ref{3dtsne}. From the t-SNE visualization, it is evident that samples belonging to the ``dog" category share close feature space with those from the ``horse" category. Similarly, the relationship between ``elephant" and ``giraffe" appears to be proximate, while ``person" and ``house" also exhibit a close association. These observed correspondences align well with the semantic representations in the word vectors, validating their consistent semantic relevance.
\subsubsection{t-SNE Visualization on Data Doamin}
\begin{figure}[htbp]
    \centering
    \includegraphics[width=\linewidth]{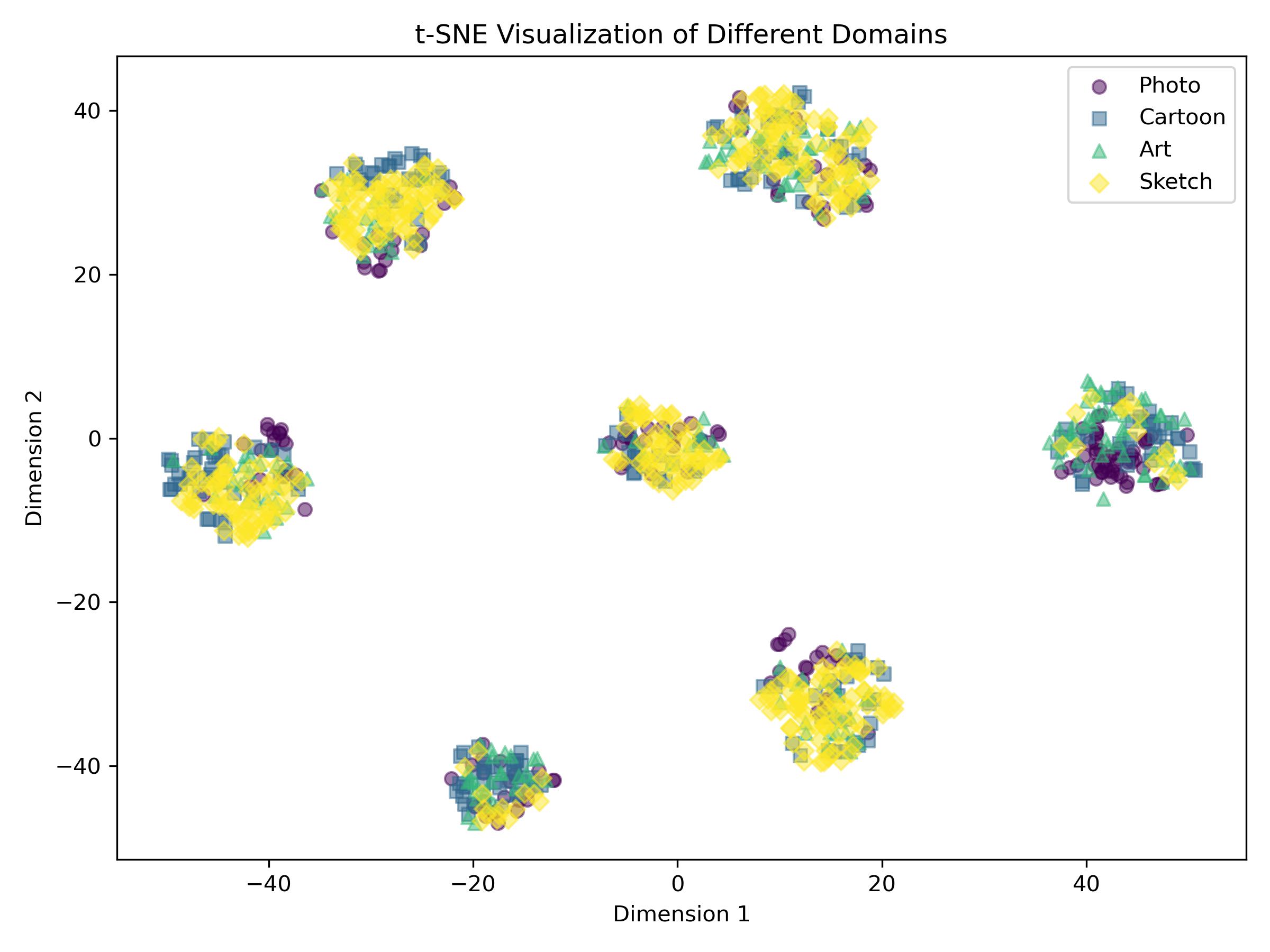}
    \caption{The distribution of the source and the target data.}
    \label{datatsne}
\end{figure}

We show the t-SNE visualization on differernt data domain features in Fig. \ref{datatsne}. It can be seen that data from the four doamins are mix up with each other according to categories.
\subsubsection{Feature Heatmap Visualization}
\begin{figure*}
	\centering
	\subfloat{
		\rotatebox{90}{\scriptsize{\small~~~~~~Caltech101}}
		\begin{minipage}[t]{0.15\linewidth}
			\includegraphics[height=1\linewidth,width=1\linewidth]{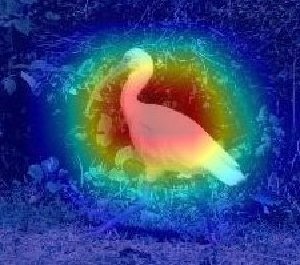}  
		\end{minipage}
		\begin{minipage}[t]{0.15 \linewidth}
			\includegraphics[height=1\linewidth,width=1\linewidth]{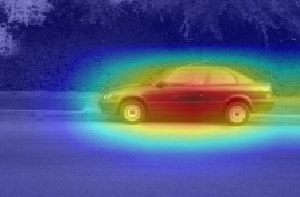}  
		\end{minipage}
		\begin{minipage}[t]{0.15\linewidth}
			\includegraphics[height=1\linewidth,width=1\linewidth]{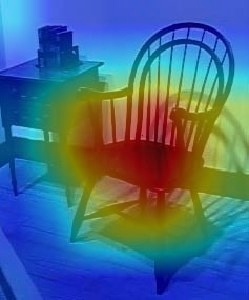}  
		\end{minipage}
		\begin{minipage}[t]{0.15\linewidth}
		\includegraphics[height=1\linewidth,width=1\linewidth]{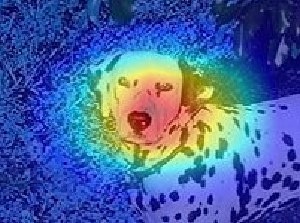}  
		\end{minipage}
		\begin{minipage}[t]{0.15\linewidth}
			\includegraphics[height=1\linewidth,width=1\linewidth]{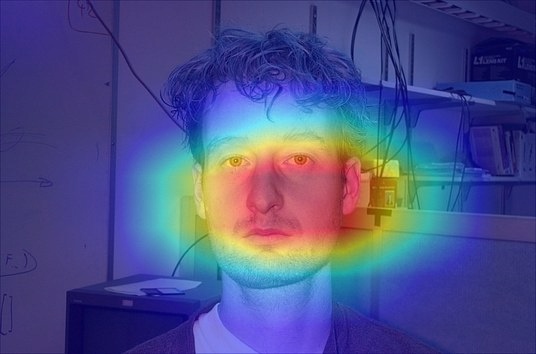}  
		\end{minipage}}
  \vspace{-2mm}
	\subfloat{
		\rotatebox{90}{\scriptsize{\small~~~~~~LabelMe}}
		\begin{minipage}[t]{0.15\linewidth}
			\includegraphics[height=1\linewidth,width=1\linewidth]{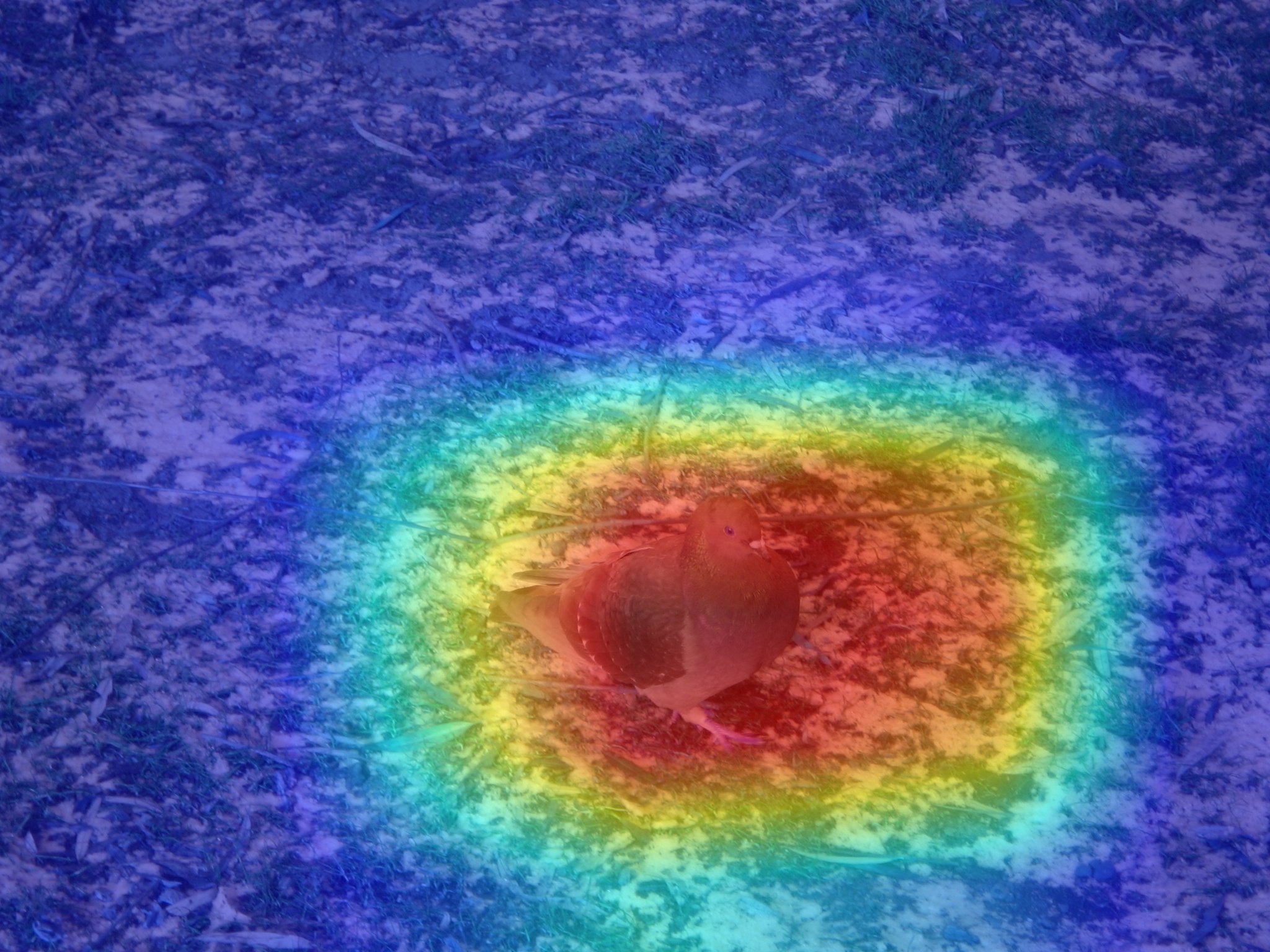}  
		\end{minipage}
		\begin{minipage}[t]{0.15\linewidth}
			\includegraphics[height=1\linewidth,width=1\linewidth]{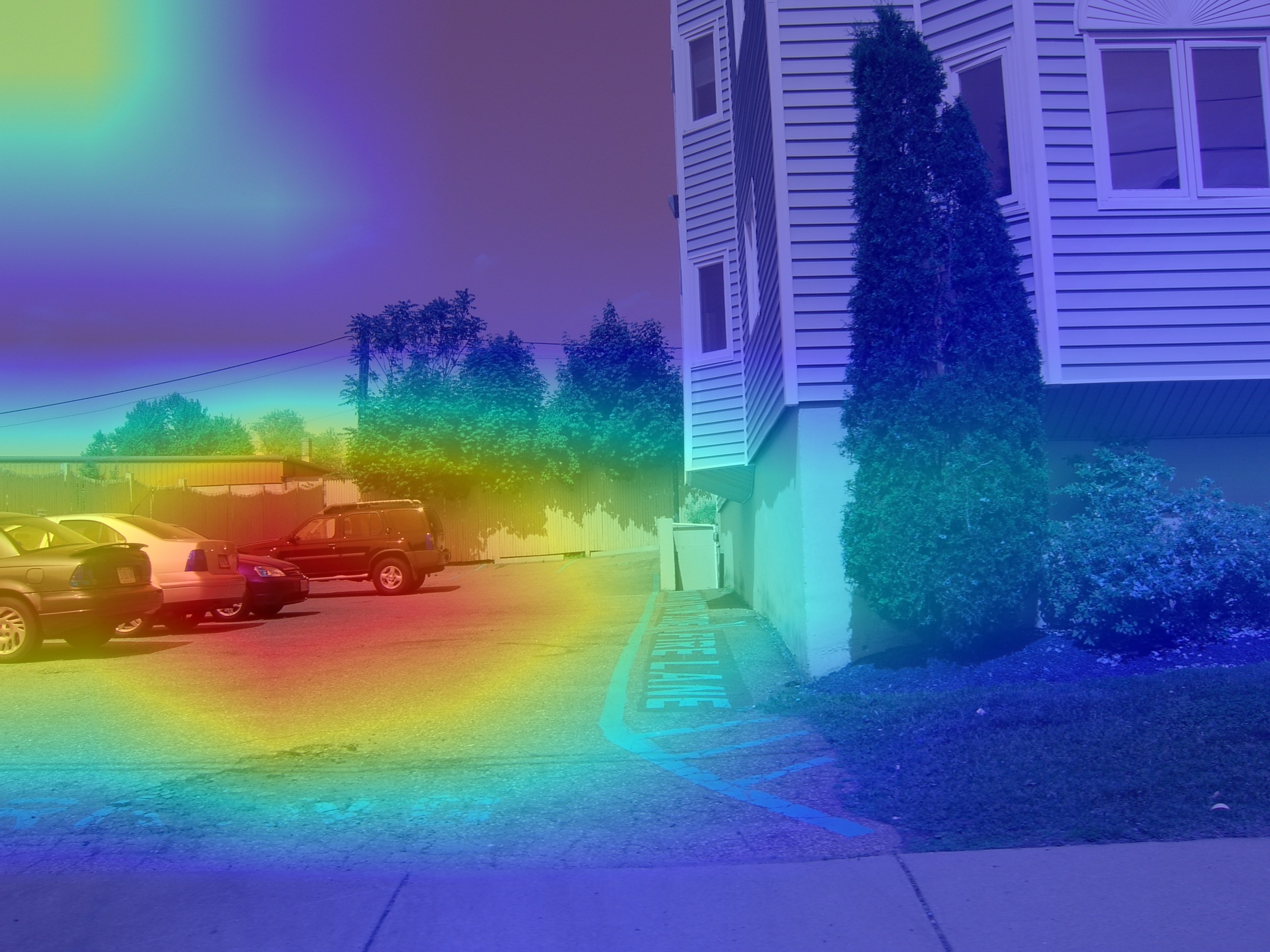}  
		\end{minipage}
		\begin{minipage}[t]{0.15\linewidth}
			\includegraphics[height=1\linewidth,width=1\linewidth]{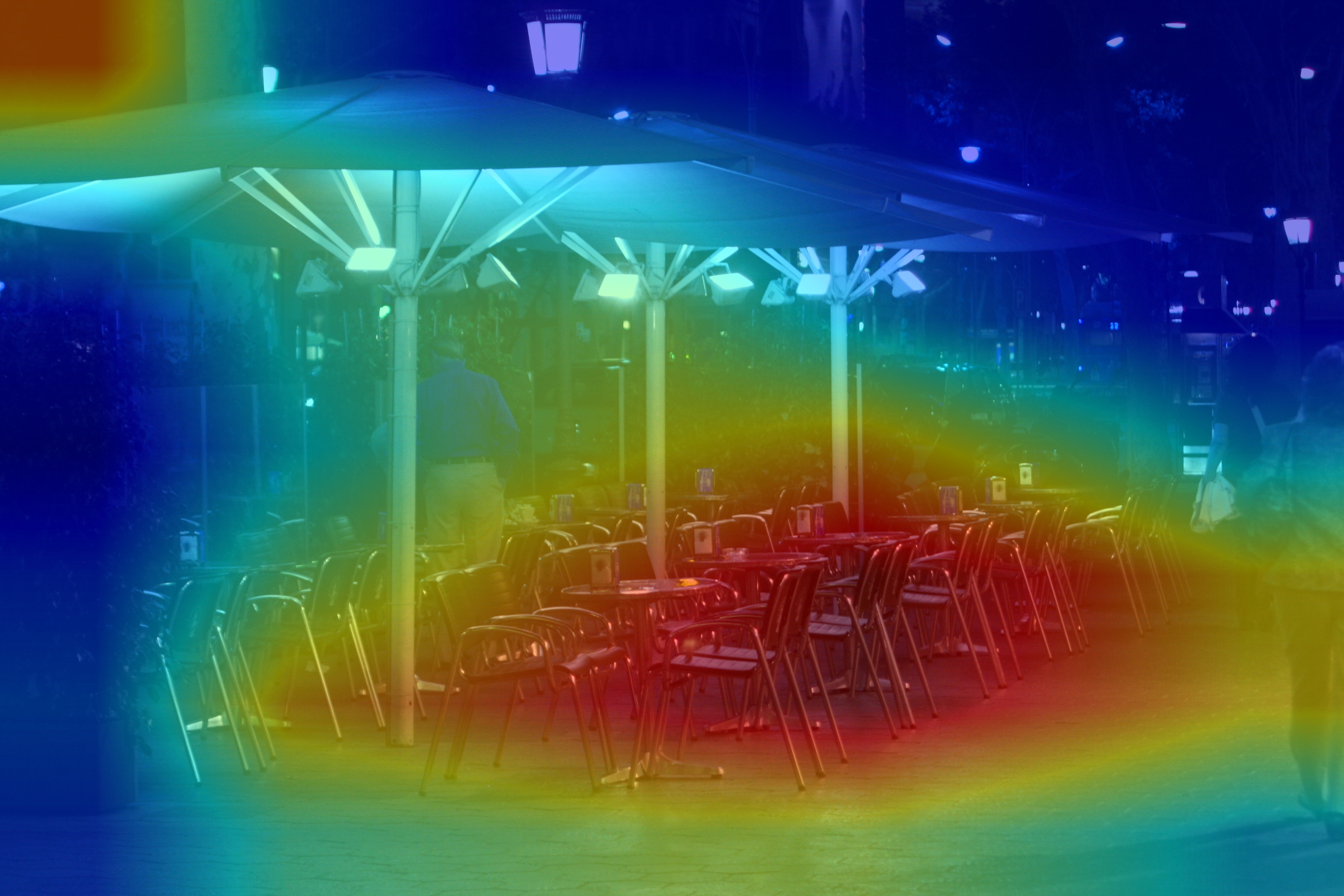}  
		\end{minipage}
		\begin{minipage}[t]{0.15\linewidth}
			\includegraphics[height=1\linewidth,width=1\linewidth]{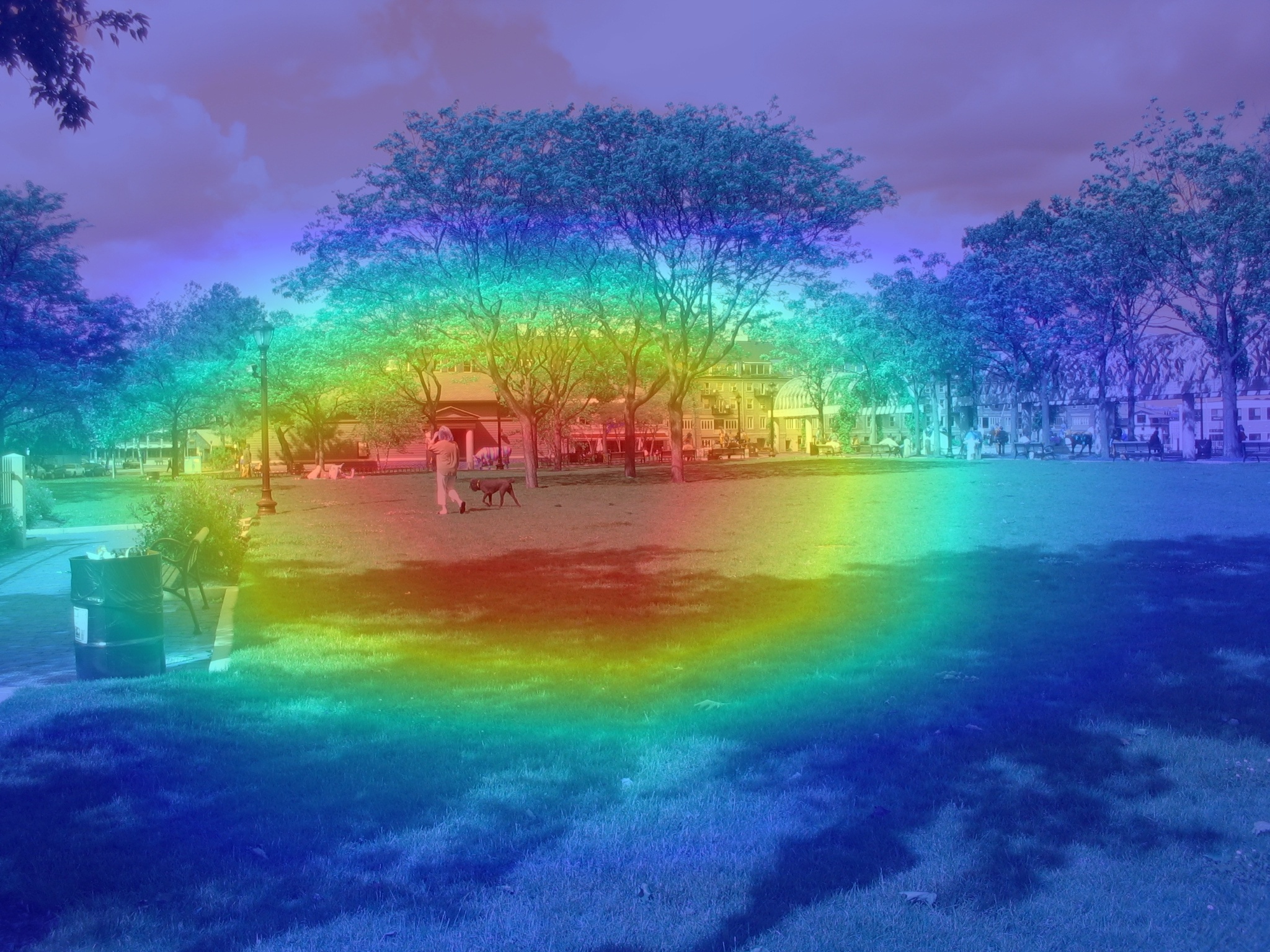}  
		\end{minipage}
		\begin{minipage}[t]{0.15\linewidth}
			\includegraphics[height=1\linewidth,width=1\linewidth]{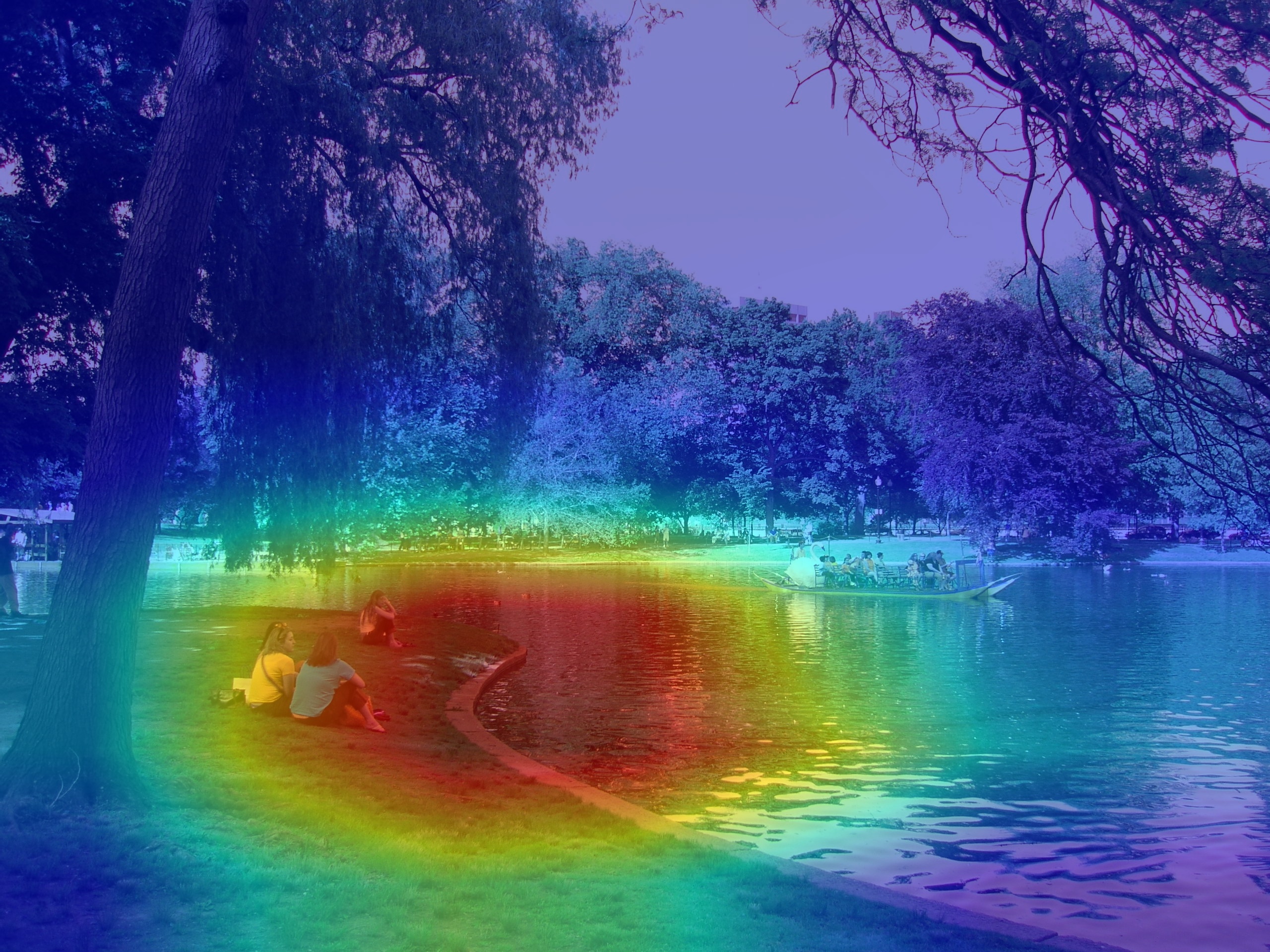}  
	\end{minipage}}
	\vspace{-2mm}
		\subfloat{
		\rotatebox{90}{\scriptsize{\small~~~~~~~SUN09}}
		\begin{minipage}[t]{0.15\linewidth}
			\includegraphics[height=1\linewidth,width=1\linewidth]{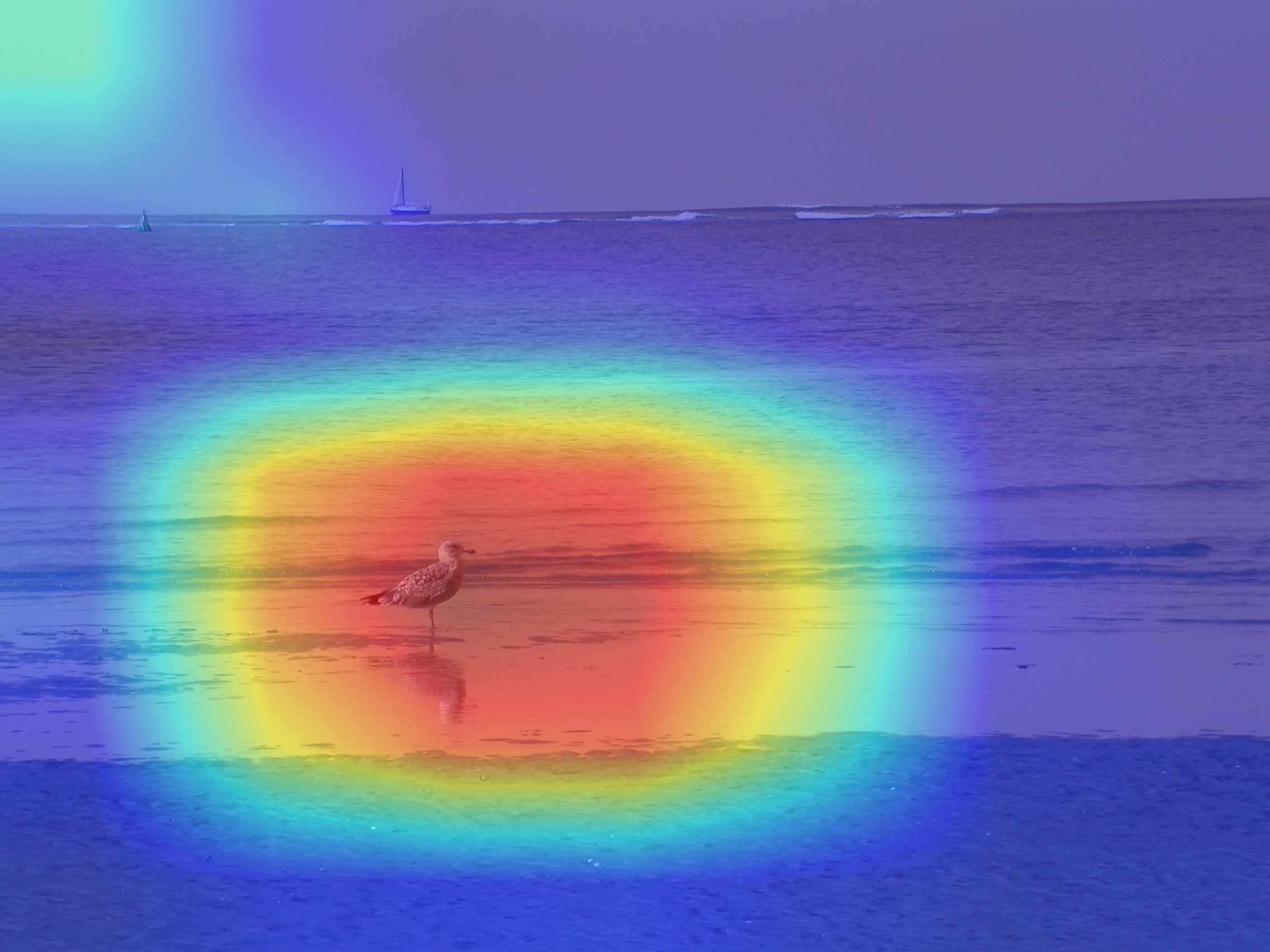}  
		\end{minipage}
		\begin{minipage}[t]{0.15\linewidth}
			\includegraphics[height=1\linewidth,width=1\linewidth]{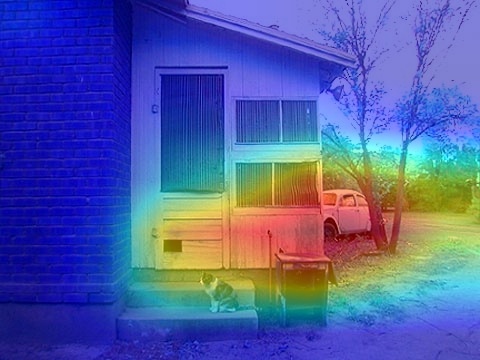}  
		\end{minipage}
		\begin{minipage}[t]{0.15\linewidth}
			\includegraphics[height=1\linewidth,width=1\linewidth]{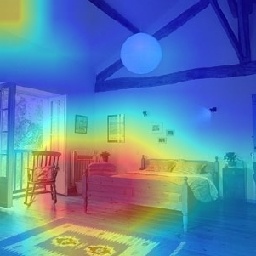}  
		\end{minipage}
		\begin{minipage}[t]{0.15\linewidth}
			\includegraphics[height=1\linewidth,width=1\linewidth]{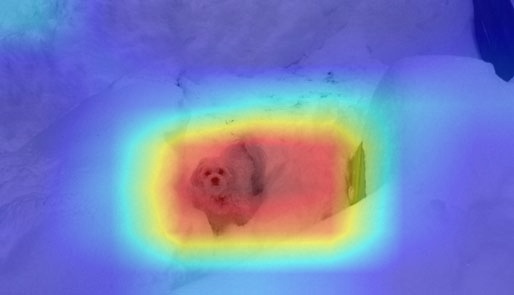}  
		\end{minipage}
		\begin{minipage}[t]{0.15\linewidth}
			\includegraphics[height=1\linewidth,width=1\linewidth]{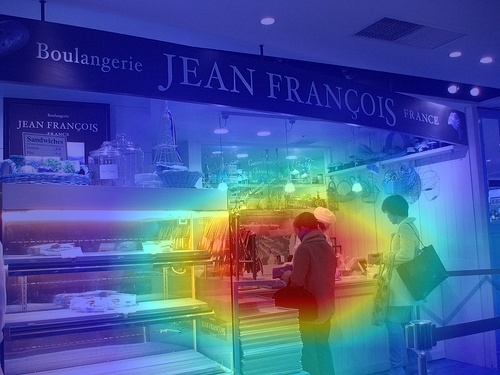}  
	\end{minipage}}
 \vspace{-2mm}
	\subfloat{
		\rotatebox{90}{\scriptsize{\small~~~~VOC2007}}
		\begin{minipage}[t]{0.15\linewidth}
			\includegraphics[height=1\linewidth,width=1\linewidth]{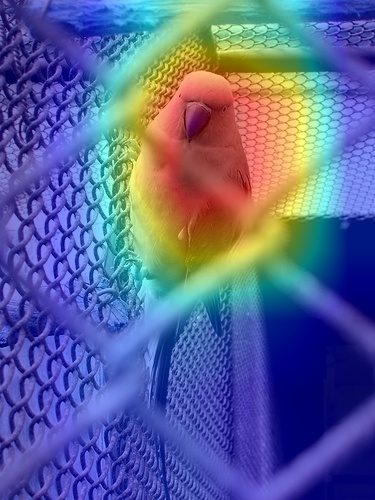}  
		\end{minipage}
		\begin{minipage}[t]{0.15\linewidth}
			\includegraphics[height=1\linewidth,width=1\linewidth]{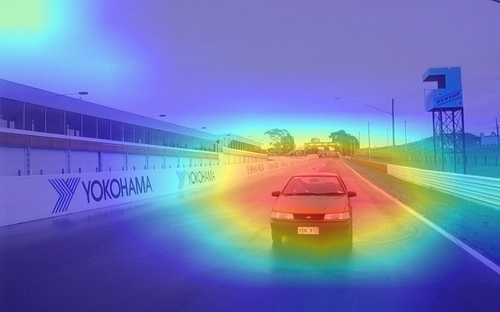}  
		\end{minipage}
		\begin{minipage}[t]{0.15\linewidth}
			\includegraphics[height=1\linewidth,width=1\linewidth]{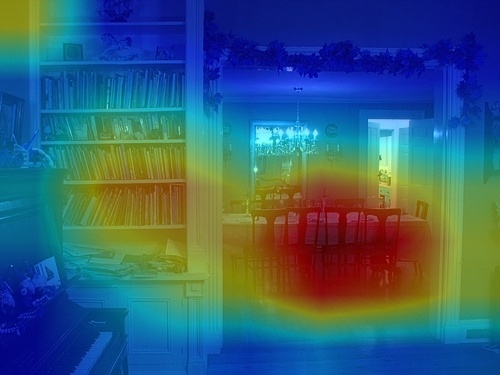}  
		\end{minipage}
		\begin{minipage}[t]{0.15\linewidth}
			\includegraphics[height=1\linewidth,width=1\linewidth]{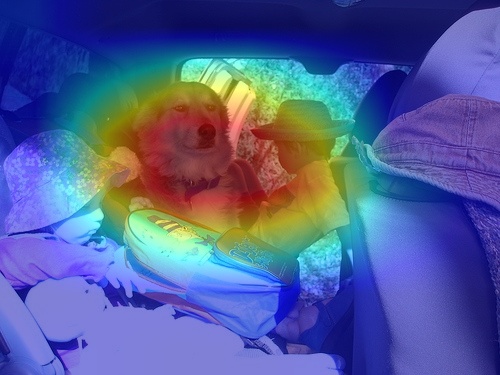}  
		\end{minipage}
		\begin{minipage}[t]{0.15\linewidth}
			\includegraphics[height=1\linewidth,width=1\linewidth]{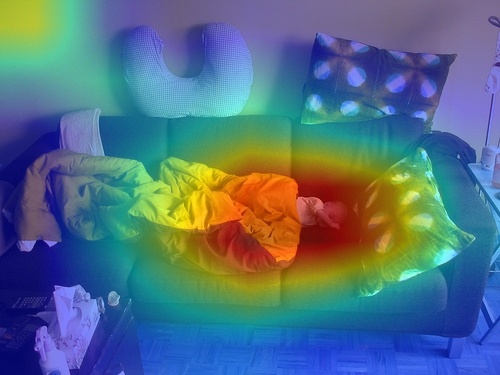}
		\end{minipage}}
		 
	\caption{The feature visualization of leave-one-domain models on VLCS with ResNet-50. The leave-one-domain is set to LabelMe. Each image is composed of the original image overlaid with its corresponding feature heatmap. In the heatmap, red corresponds to positions with high feature values, while blue corresponds to positions with lower feature values. }
	\label{heatmap}
 % \vspace{-3mm}
\end{figure*}

%The heatmap results on VLCS dataset with ResNet 50 are shown in Figure \ref{heatmap}. The figure displays the feature heatmaps of images from four domains and five categories respectively. In each feature heatmap, red highlights the regions of particular interest to the network. It can be seen that the regions of particular interest to the network corresponds to discriminative class-specific features.
The heatmap results for the ResNet 50 model on the VLCS dataset are presented in Fig. \ref{heatmap}. This figure provides an overview of the feature heatmaps for images spanning four distinct domains and five diverse categories. 
%Within each feature heatmap, the prominence of red hues signifies the network's heightened attention to specific areas of the image. 
This emphasis aligns with the presence of class-specific features that hold discriminative significance. The discernible correlation between the network's focus and these distinctive features underscores the model's ability to recognize and highlight meaningful patterns.
\subsubsection{Ablation Study on Batchsize}\label{abla_batchsize}
\begin{figure}
    \centering
    \includegraphics[width=0.5\textwidth]{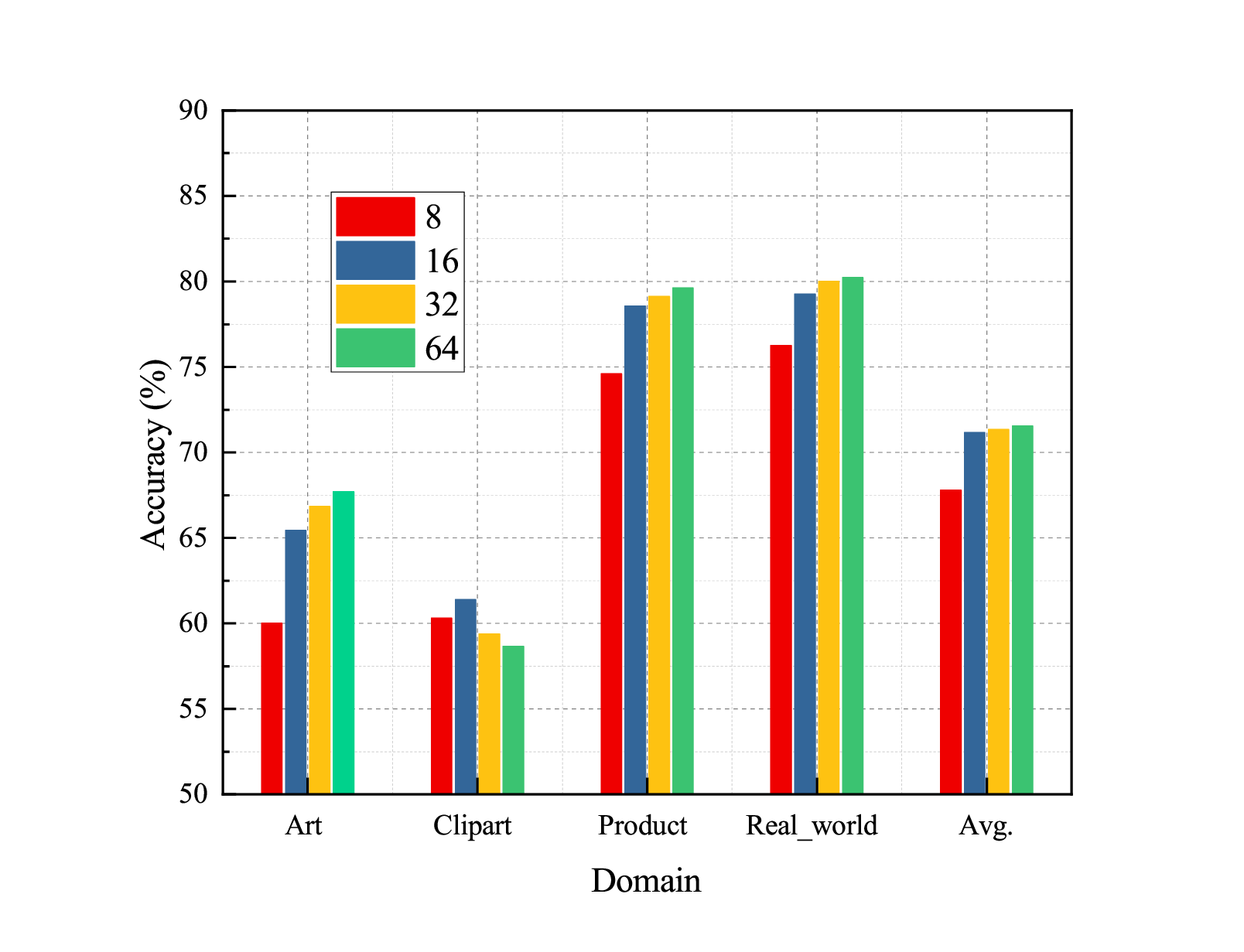}
    \caption{The leave-one-domain results on OfficeHome with ResNet-50 under different batchsize.}
    \label{batchsize}
\end{figure}

We conduct experiments with different batch sizes as shown in Fig. \ref{batchsize}. Specifically, experiments are conducted with batch sizes of 8, 16, 32, and 64. Due to limitations in GPU memory, we are unable to test larger batch sizes. The experimental results indicate a continuous improvement in overall performance as the batch size increases.  This can be attributed to the fact that with larger batch sizes, the low-rank approximate module for intra-class consistency constraints can identify more shared patterns among samples from different domains but of the same class. Consequently, this leads to a reduction in inter-domain disparities, an increase in intra-class differences, and an enhancement in the robustness and generalization capabilities of the model. The main body of the Clipart dataset clearly has minimal background interference. The reason for the decrease in performance on Clipart domain may be that when the batch size increases, the background information from images in other domains may interfere with the features of the Clipart domain images.
% \begin{table}[htbp]
% 	\caption{The leave-one-domain results on OfficeHome with ResNet-50 under different batchsize.}
% 	\centering
% 	%\resizebox{0.485\textwidth}{!}{
% 		\begin{tabular}{@{}cccccc@{}}
% 			\toprule
% 			Batchsize               & Art   & Clipart & Product & Real\_world                & Avg.  \\ \midrule
% 			\multicolumn{1}{c|}{8}  & 59.99 & 60.30    & 74.59   & \multicolumn{1}{c|}{76.25} & 67.78 \\
% 			\multicolumn{1}{c|}{16} & 65.43 & 61.37   & 78.55    &    \multicolumn{1}{c|}{79.25} & 71.15  \\
% 			\multicolumn{1}{c|}{32} & 66.83 & 59.36   & 79.12   & \multicolumn{1}{c|}{80.01} & 71.33 \\
% 			\multicolumn{1}{c|}{64} & 67.70  & 58.64   & 79.61   & \multicolumn{1}{c|}{80.22} & 71.54 \\ \bottomrule
% 	\end{tabular}
%  %}
% 	\label{batchsize}
% 	%\vspace{-5mm}
% \end{table}
\subsubsection{Ablation Study on Word Vector Model}\label{abla_wordvec}
To explore the influence of word vector models on performance, we conduct tests using open-source word vector models, including GloVe and Word2Vec as shown in TABLE \ref{wordvector}. The experimental results reveal that ``glove.42B.300d" outperforms the others. This can be attributed to the fact that this model is based on a large corpus and employs high-dimensional representations for each word, enabling fine-grained characterization.
\begin{table}[!ht]
	\centering
	\caption{The leave-one-domain results by using different word vector model on VLCS with ResNet-50. }
	%\resizebox{0.485\textwidth}{!}{
	\begin{tabular}{@{}l|ccccc@{}}
		\toprule
		Word Vector Model              & C & L & S & V & Avg.  \\ \midrule
		glove.6B.50d                   & 98.80      & 67.78   & 77.12 & 78.41   & 80.53 \\
		glove.6B.100d                  & 99.29      & 68.60   & 74.62 & 78.29   & 80.20 \\
		glove.6B.200d                  & 99.08      & 66.98   & 76.36 & 78.44   & 80.22 \\
		glove.6B.300d                  & 98.94      & 67.51   & 76.64 & 78.82   & 80.48 \\
		glove.42B.300d                 & 99.01      & 67.77   & 77.06 & 79.06   & 80.73 \\
		glove.twitter.27B.25d          & 98.80      & 67.24   & 75.62 & 79.59   & 80.31 \\
		\begin{tabular}[l]{@{}l@{}}GoogleNews-vectors-\\ negative300\end{tabular} & 98.73      & 67.28   & 75.93 & 78.70   & 80.16 \\ \bottomrule
	\end{tabular}
 %}
\label{wordvector}
%\vspace{-5mm}
\end{table}
% \usepackage{booktabs}
\subsubsection{Ablation Study on CLIP Prtrained Model}\label{abla_clip}
To compare the impact of CLIP pre-trained models on domain prompt encoding, we conduct tests on different pre-trained models, and the results are shown in TABLE \ref{clip}. The experimental findings indicate that RN50x16 demonstrates the best overall performance, outperforming other models.
\begin{table}[htbp]
	\caption{The leave-one-domain results by using different CLIP pre-trained models on VLCS with ResNet-50.}
	\resizebox{0.485\textwidth}{!}{
	\begin{tabular}{@{}c|ccccc@{}}
		\toprule
		CLIP pre-trained model & C     & L     & S     & V     & Avg.  \\ \midrule
		RN50                  & 98.94 & 66.38 & 76.75 & 78.02 & 80.02 \\
		RN101                 & 98.80 & 67.78 & 77.12 & 78.41 & 80.53 \\
		RN50x4                & 98.87 & 67.85 & 76.20 & 77.87 & 80.20 \\
		RN50x16               & 98.73 & 70.26 & 76.23 & 78.23 & 80.86 \\
		ViT-B/32              & 98.94 & 67.39 & 75.26 & 78.52 & 80.03 \\
		ViT-B/16              & 99.08 & 67.21 & 74.89 & 80.27 & 80.36 \\
		ViT-L/14              & 99.15 & 68.83 & 76.45 & 76.78 & 80.30 \\
		ViT-L/14@336px        & 99.22 & 68.11 & 75.50 & 79.32 & 80.54 \\ \bottomrule
	\end{tabular}}
\label{clip}
\end{table} 

\section{Conclusion}
In this paper, we combine language space and image space, using semantic space as the bridge domain, to explore the generalization ability of models in unknown target domains. In the language space,  relative relationships between categories are explored through word vectors. In image space, the common pattern among sample features of the same category is sought by mapping the feature matrix to the low-rank subspace. In multimodal space, domain-related image features are removed through CLIP-based text domain prompts. Given the current state of incompleteness in language models, the experiment results have achieved competitive performance. 
\section{Limitation and Discussion}
Though our algorithm has good performance on the dataset, the proposed three modules still have limitations. Firstly, The domain information orthogonal decoupling module is implemented based on the semantic prompt of the CLIP pre-trained model. This requires that the semantic prompt be meaningful, that is, the domain features are clear, such as rain, fog, cartoons, sketches, etc. However, for datasets with unclear domain information, such as Digits or VLCS, the effect is reduced. Secondly, word vector interclass constraint module is limited by the completeness of the word vector model. When the category words in the dataset cannot be retrieved in the word vector model, the semantics of the substitute words may be different from that of the original category words, which reduces the accuracy of the interclass relationship. At last, the low-rank decomposition intra-class constraint module is affected by batch size. Since the purpose of this module is to find common feature patterns of the same category samples, larger batch sizes have been shown to achieve better results.

With the ongoing advancements in natural language processing, it is expected that language models will offer more robust supervision, thereby enhancing domain generalization capabilities. In addition, image basic units with semantic properties are also worth exploring. Based on the above, with the development of large language models and the further improvement of GPU memory, the performance of the proposed method has room for further improvement.
\bibliographystyle{cas-model2-names}

\bibliography{cas-dc-template}

%% Biography
%% Here goes the biography details.
%% Here goes the biography details.

\end{document}